\documentclass[pdflatex,sn-mathphys-num,iicol]{sn-jnl}

\usepackage{graphicx}%
\usepackage{multirow}%
\usepackage{amsmath,amssymb,amsfonts}%
\usepackage{amsthm}%
\usepackage{cleveref}
\usepackage{mathrsfs}%
\usepackage[title]{appendix}%
\usepackage{xcolor}%
\usepackage{textcomp}%
\usepackage{manyfoot}%
\usepackage{booktabs}%
\usepackage{algorithm}%
\usepackage{algorithmicx}%
\usepackage{algpseudocode}%
\usepackage{tikz}
\usepackage{subcaption}
\usepackage{pgfplots}
\usepackage{listings}%
\usepackage{hyperref}
\theoremstyle{thmstyleone}%
\theoremstyle{thmstyletwo}%

\theoremstyle{thmstylethree}%

\begin{document}

\title[Robust Robotic Exploration and Mapping Using Generative Occupancy Map Synthesis]{Robust Robotic Exploration and Mapping Using Generative Occupancy Map Synthesis}

% \title[Generative Occupancy Mapping for
% Enhanced Robotic Exploration]{Generative Occupancy Mapping for
% Enhanced Robotic Exploration}


\author[]{\fnm{Lorin} \sur{Achey}}\email{lorin.achey@colorado.edu}

\author[]{\fnm{Alec} \sur{Reed}}\email{alec.reed@colorado.edu}

\author[]{\fnm{Brendan} \sur{Crowe}}\email{brendan.crowe@colorado.edu}

\author[]{\fnm{Bradley} \sur{Hayes}}\email{bradley.hayes@colorado.edu}

\author*[]{\fnm{Christoffer} \sur{Heckman}}\email{christoffer.heckman@colorado.edu}

\affil[]{\orgdiv{Department of Computer Science}, \orgname{University of Colorado Boulder}, \orgaddress{\street{1111 Engineering Dr}, \city{Boulder}, \postcode{80309}, \state{Colorado}, \country{United States}}}


\abstract{We present a novel approach for enhancing robotic exploration by using generative occupancy mapping. We implement SceneSense, a diffusion model designed and trained for predicting 3D occupancy maps given partial observations. Our proposed approach probabilistically fuses these predictions into a running occupancy map in real-time, resulting in significant improvements in map quality and traversability. We deploy SceneSense on a quadruped robot and validate its performance with real-world experiments to demonstrate the effectiveness of the model. In these experiments we show that occupancy maps enhanced with SceneSense predictions better estimate the distribution of our fully observed ground truth data ($24.44\%$ FID improvement around the robot and $75.59\%$ improvement at range). We additionally show that integrating SceneSense enhanced maps into our robotic exploration stack as a ``drop-in'' map improvement, utilizing an existing off-the-shelf planner, results in improvements in robustness and traversability time. Finally, we show results of full exploration evaluations with our proposed system in two dissimilar environments and find that locally enhanced maps provide more consistent exploration results than maps constructed only from direct sensor measurements.}

%%================================%%
%% Sample for structured abstract %%
%%================================%%

% DELETED




\maketitle

\section{Introduction}\label{sec1}

Efficient and reliable robotic exploration is fundamental to the practical deployment of robotic systems for use in real-world scenarios. In order to intelligently explore, an autonomous agent must have a useful representation of its surroundings and be able to localize within the environment. Robotic agents deployed in search and rescue \cite{schwarz2017nimbro, bernard2011autonomous} or disaster response missions \cite{kawatsuma2012emergency, park2017disaster, kruijff2021germanemergencyresponse} rarely have the environment mapped a priori. The robotic agent must operate cautiously as it builds the map from direct sensor observations. This results in slower, less efficient exploration as the agent can only operate over terrain in the sensor field-of-view and large portions of the space may be obscured from that view.

This lack of prior scene information means agents are constantly operating on incomplete information. In addition to limited field-of-view for mapping, there are sensor noise issues, unpredictable dynamic obstacles, obstructions of key information, and visual degradation that an agent may need to reason over to complete a given task. While human agents can filter out noise and fill in gaps in observations by drawing on prior experiences, robots are generally limited to using direct observations without the benefits of prior knowledge or intuition. This results in robotic agents that only take action based on direct observations, a process that further reduces exploration efficiency because of time consuming replanning when informative sensor updates occur.

In related fields, such as self-driving cars, researchers attempt to eliminate the mapping uncertainty using pre-compiled high-definition (HD) maps \cite{ma2019exploiting}. This approach provides a good method of localization at the global level, but there still exists uncertainty and missing information locally. Furthermore, this approach scales poorly, as new HD maps are required for any new operating environment. As discussed previously there are many applications where collating an HD map before navigation is not possible, such as in search and rescue scenarios \cite{chung2023into} or extra-terrestrial exploration \cite{Verma_2024_perseverance_mars}.

With recent breakthroughs in generative AI, neural networks are able to generate convincing human-like results in various modalities, such as text \cite{touvron2023llama}, images \cite{rombach2022high, kapelyukh2023DALLEBOT}, and audio \cite{dhariwal2020jukebox}. We leverage these models to enable robots to make common-sense inference of geometry that is occluded from view. Our model enables a robotic agent to infer geometry through occupancy prediction beyond an obstruction so that the agent can plan a trajectory beyond the field-of-view similar to the way a human would navigate occlusions using intuition. 

SceneSense was first introduced in \cite{reed2024scenesense}, where a diffusion-based model was used to generate voxelized predictions of occupancy. Subsequent work improved the efficiency of SceneSense, notably achieving faster training and inference times \cite{reed2024online}. These works established that diffusion models can be used for occupancy prediction but left open questions regarding real-time deployment and map level reasoning.

In this work, we extend the SceneSense framework in three ways. First, we demonstrate real-time occupancy generation on a quadruped platform using a diffusion network. Second, we integrate a probabilistic update rule that fuses diffusion based predictions with an evolving observed map, allowing predictions to accumulate and refine over time. Third, we show that SceneSense can generate realistic occupancy predictions across the full map rather than being restricted to regions local to the robot. This broader predictive capability leads to enhanced global maps which can simplify exploration in real world deployments.

We validate our approach through field experiments with a quadruped platform. Specifically, we (i) compare robot-centric and frontier-centric exploration performance, (ii) evaluate one-shot versus probabilistic map merging, and (iii) demonstrate successful navigation in challenging scenarios that historically degrade robotic perception (e.g. transparent structures, narrow hallways). Finally, we provide detailed reporting of system specifications, compute requirements, training times, experimental design, and system limitations. Code for reproducing our results will be made available at [URL-added-post-publication].

\section{Related Works}\label{sec2}

% Sample body text. Sample body text. Sample body text. Sample body text. Sample body text. Sample body text. Sample body text. Sample body text.

% \section{This is an example for first level head---section head}\label{sec3}

% \subsection{This is an example for second level head---subsection head}\label{subsec2}

% \subsubsection{This is an example for third level head---subsubsection head}\label{subsubsec2}

% Sample body text. Sample body text. Sample body text. Sample body text. Sample body text. Sample body text. Sample body text. Sample body text. 

\subsection{Scene Exploration}

An autonomous agent must have a useful representation of its surroundings and be able to localize within the environment to efficiently and intelligently explore. A proven strategy for mapping complex environments for use with autonomous systems is 3D volumetric mapping \cite{ahmad20213dunstructured, ahmad2021reactiveobstacleavoidance}. OctoMap \cite{hornung13octomap} is a popular framework for 3D mapping due to its memory efficiency and wide adoption in the robotics community \cite{dang2020graph}. In the DARPA Subterranean (SubT) Challenge \cite{chung2023into}, Team MARBLE \cite{biggie2023flexible} utilized OctoMap on a Boston Dynamics Spot robot and finished 3rd overall.

\begin{figure*}
    \centering
    \includegraphics[width=1.0\linewidth]{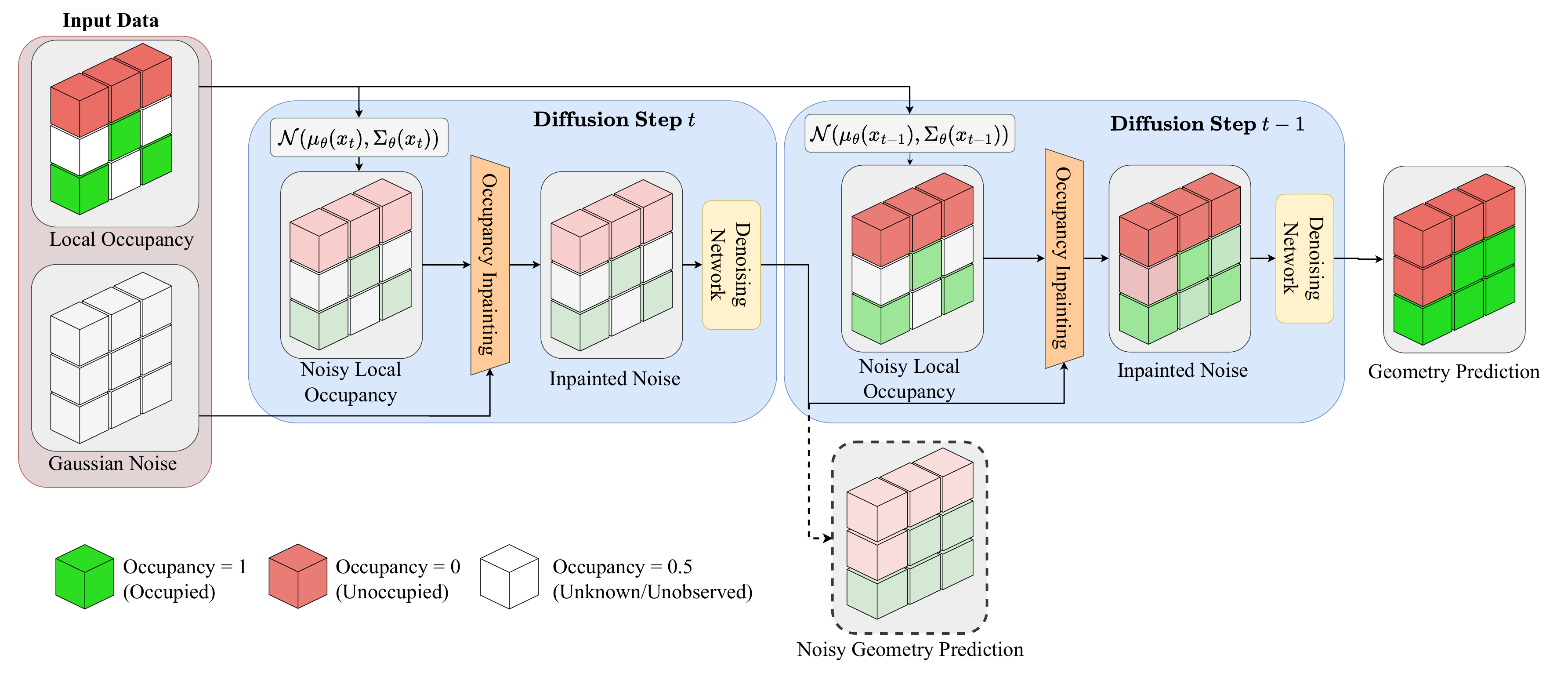}
    \caption{The reverse diffusion process takes the local occupancy information and the Gaussian noise of the area to be diffused over. Noise commensurate with the current diffusion step is added to the local occupancy information, which includes occupied (green) and observed unoccupied (red) data. The result is inpainted into the noisy local occupancy prediction. The inpainted noise data is provided to the denoising network which generates a new noisy geometry prediction at $t-1$.  This process is repeated as the starting noise $x_T$ is iteratively denoised to $x_0$ which is the final geometry prediction from the framework. This process is further detailed in Algorithm \ref{alg:unconditional_diffusion}.}
    \label{fig:diffusion_framework}
\end{figure*}

Other works \cite{yamauchi1997frontier, yamauchi1998mobile, makarenko2002experiment, gonzalez2002navigation} have shown considerable success using frontier-based exploration strategies. In \cite{freda2005frontier}, a frontier-based approach is utilized to demonstrate the efficacy of biasing towards unseen areas of a map for more effective mobile robot exploration. Further efficiency improvements to the frontier based exploration strategy are introduced in \cite{keidar2014efficient}. In \cite{gonzalez2002navigation}, a frontier based method is proposed for safe and efficient map-building by defining a ``safe region" around the robot, allowing it to explore and gather new information while ensuring obstacle-free navigation, although it is limited to mapping at a fixed height, making it unsuitable for more complex environments.

In \cite{CERBERUStranzatto2022team}, a volumetric map is incrementally generated from depth sensors and a graph-based planner \cite{dang2020graph} reasons over that map to determine if the terrain is traversable for each vertex of a four-legged robot. This graph traversal requires considerable computation time for determining the status of voxels beneath and around the robot platform. Additionally, this method does not address the issues with path planning if there are obstacles or geometric occlusions in the sensor field of view.

Despite the improvements in mapping and frontier-based exploration strategies, Team MARBLE \cite{biggie2023flexible} reported that there are still challenges with efficient exploration. During the SubT challenge, exploring robots frequently wasted valuable time attempting to explore in previously seen areas rather than extending their trajectories beyond their current frontiers \cite{biggie2023flexible}. Influxes of new information from onboard sensors that came when turning corners or exploring new rooms caused the robots to pause for extended periods of time for re-planning, slowing the pace of exploration considerably \cite{biggie2023flexible}. The challenge outcomes demonstrated the utility of 3D volumetric mapping using frontier-based exploration for discovery, but also elucidated the need for increased efficiency in the planning and exploration process for time critical missions. Our novel approach improves the efficiency of exploration by reducing ambiguity that leads to long re-planning delays for the robots.

\subsection{Occupancy Prediction}

\citet{wang2021learning3doccupancyprediction} attempts to address challenges of autonomous navigation in environments with occlusions by utilizing a deep learning-based approach to predict the occupancy distribution of portions of the environment. Their method, the Occupancy Prediction Network (OPNet), predicts obstacle distributions, utilizing a self-supervised learning technique that generates training data through simulated navigation trajectories with simulated sensor noise. This approach relies on removing portions of the data from the Matterport3D \cite{chang2017matterport3d} dataset to train the neural network which biases the model to predict occupancy only for the types of occlusions removed from the original dataset. This approach does not scale well to large unseen sections of an environment, is limited to static updates, and does not generalize due to its reliance on a specially curated dataset.

More recently, \citet{Huang_2024_CVPR} introduced SelfOcc, a self-supervised method for 3D occupancy prediction using video sequences. SelfOcc converts 2D images into 3D representations with deformable attention layers and uses signed distance fields (SDF) for regularization and occupancy boundary determination. Although SelfOcc performs well with surrounding cameras, it cannot predict occupancy beyond the camera view due to the limitations of SDF in handling occluded geometry.

\subsection{Scene Synthesis}

Diffusion models \cite{sohl2015deep, ho2020denoising}, are a popular and promising technique which have demonstrated impressive results in image \cite{rombach2022high}, video \cite{harvey2022flexible}, and natural language \cite{huang2022prodiff}. Based on these successes, diffusion models are being extended to 3D scene and shape generation. Recent work \cite{luo2021diffusionpointcloudgeneration} demonstrates the use of diffusion models for 3D point cloud generation for simple shapes and objects (e.g. tables, chairs).  \citet{kim2023scenegendiffusionmodels} shows successful 3D shape generation from 2D content such as images, and \citet{vahdat2022lion} demonstrates similar 3D shape generation but using point cloud datasets rather than images. In LegoNet \cite{yu2022legonet}, diffusion models are used to rearrange objects in a 3D scene. In DiffuScene \cite{tang2023diffuscene}, a denoising diffusion model is used with text conditioning to generate 3D indoor scenes from sets of unordered object attributes. Unlike these previous works which primarily focus on generating simple shapes, rearranging objects, or creating indoor scenes, our approach leverages diffusion models to fuse generated terrain with the local robot field of view, thereby bridging the gap between 3D scene generation and practical robotics applications.

\subsection{Generative AI in Robotics}

The body of work devoted to applications of generative AI in robotics is steadily growing. \citet{yuan2023hierarchicalgenmodeling} demonstrates that hierarchical generative modeling, inspired by human motor control, can enable autonomous robots to effectively perform complex tasks with robust performance even in challenging conditions. DALL-E-Bot \cite{kapelyukh2023DALLEBOT}, uses web-scale diffusion models for generating an image from a text prompt which the robot utilizes to rearrange real objects in accordance with the image. Diffusion models were also used in \cite{carvalho2023planningdiffusionmodels} to improve robot motion planning by learning priors on trajectory distributions from previously successful plans, a methodology which shows good generalization capabilities in environments with previously unseen obstacles. More recently, diffusion models were shown to successfully generate terrain predictions behind occluded geometries in indoor environments using a single RGB sensor mounted on a mobile robot platform \cite{reed2024scenesense}. We build on this work, extending the practical applications of diffusion models by demonstrating a novel method for occupancy prediction which shows significant efficiency gains during exploration.

%%%%%%%%%%%%%%%%%%%%%%%%%%%%%%%%%%%%%
% PROBLEM STATEMENT AND PRELIMINARIES
%%%%%%%%%%%%%%%%%%%%%%%%%%%%%%%%%%%%%

\section{Problem Statement and Preliminaries}\label{sec3}
% At the highest level, the problem that is addressed in this paper is one of autonomous robotic exploration. However we also address the problem of frontier identification as well as the problem of dense occupancy prediction

The goal of this work is to create and evaluate a method for dense occupancy prediction that demonstrably enhances a robotic platform’s ability to navigate, as measured by reducing the time required to fully explore its environment. The ability to generate occupancy predictions beyond occlusions and at range from a mobile platform introduces the subsequent challenge of how to identify and select frontier points for exploration in these generated occupancy spaces. We present an approach to frontier identification and selection for exploration as part of this work.

\subsection{Autonomous Robotic Exploration}
Let $\mathbb{M}$ denote the current occupancy map, constructed from onboard sensor measurements $\mathbb{S}$ and odometry $\mathbb{O}$. The map consists of voxels $m \in \mathbb{M}$, where each voxel is categorized as free ($m \in \mathbb{M}_{free}$), occupied ($m \in \mathbb{M}_{occupied}$), or unknown ($m \in \mathbb{M}_{unknown}$). During robotic exploration we seek to recover the complete map $\mathbb{M}_{comp}$ of a target region $\mathbb{M}_{target}$ from successive measurements from the onboard sensors. While accurate completion and alignment of $\mathbb{M}_{comp}$ with $\mathbb{M}_{target}$ is the primary goal, secondary goals include minimizing exploration time and maximizing volumetric gain per second, provided safety is not compromised.

\subsection{Frontier Identification and Evaluation}
 We identify and evaluate frontiers of interest in $\mathbb{M}$ that can enhance the robot's understanding of the environment during exploration. In general, ``interesting'' frontiers will maximize the number of unknown voxels available for occupancy prediction while considering common exploration metrics such as directionality, distance from target, and reachability \cite{biggie2023flexible, dang2020graph}. These frontiers guide the robot's exploration.
 
 \subsection{Dense Occupancy Prediction}
Dense occupancy prediction estimates the probability of occupancy for each voxel $m \in \mathbb{M}$, producing a continuous value in $[0,1]$ where 0 indicates free and 1 indicates occupied. These predictions are used to hypothesize structure beyond the currently observed map generated by the robots sensor measurements.

 \subsection{Forward Diffusion}
Let $x_0$ denote an occupancy subgrid ($x \subset \mathbb{M}$), sampled from the data distribution $q(x_0)$. By sampling from the data distribution $x_0 \sim q(x_0)$, the forward diffusion process is defined as a Markov chain of variables $x_1, ...,  x_T$ that iteratively adds Gaussian noise to the sample. A diffusion step at time $t$ in this chain is defined as:
\begin{equation}\label{eq:mark_step}
    q(x_t|x_{t-1}) = \mathcal{N}(x_t;\sqrt{1- \beta_t}x_{t-1},\ \beta_tI),
\end{equation}
where t is the time step $t \in [1,T]$, $\beta_t$ is the variance schedule $0 \leq \beta_t \leq 1$ and $I$ is the identity matrix. The joint distribution of the full diffusion process is then the product of the diffusion step defined in \cref{eq:mark_step}:
\begin{equation}
    q(x_{1:T}|x_{0}) = \prod_{t=1}^{T} q(x_t|x_{t-1}).
\end{equation}
Conveniently, we can apply the reparameterization trick to directly sample $x_t$ given $x_0$ using the conditional distribution:
\begin{equation} \label{eq:add_noise}
    q(x_t|x_{0}) = \mathcal{N}(x_t;\sqrt{\overline{\alpha}_t}x_{0}, (1-\overline{\alpha}_t)\mathcal{I}),
\end{equation}
where $x_t = \sqrt{\overline{\alpha}_t}x_0 + \sqrt{1 - \overline{\alpha}_t}\epsilon$ where $\alpha_t := 1 - \beta_t$, $\overline{\alpha}_t := \prod^t_{r=1}\alpha_s$, and $\epsilon$ is the noise used to corrupt $x_t$.

\subsection{Reverse Diffusion}
Reverse diffusion defines a generative process that inverts the forward noising procedure. Specifically, it is modeled as a Markov chain of learned Gaussian transitions, parameterized by a neural network with learnable parameters $\theta$:
\begin{equation}
    p_\theta(x_{t-1}|x_t) := \mathcal{N}(x_{t-1}; \mu_\theta (x_t , t) , \Sigma_\theta (x_t,t)),
\end{equation}
where $\mu_\theta(x_t, t)$ and $\Sigma_\theta(x_t, t)$ are the predicted mean and covariance of the Gaussian distribution over $x_{t-1}$. The parameters $\theta$ correspond to the SceneSense diffusion model, trained to recover occupancy maps from noisy inputs.

Given the initial state of a noisy occupancy map from a standard multivariate Gaussian distribution $x_t \sim \mathcal{N}(0,I)$, the reverse diffusion process iteratively predicts $x_{t-1}$ at each time step $t$ until reaching the final state $x_0$ which is the goal occupancy map.  Similar to the Markov chain defined forward diffusion process the joint distribution of the reverse diffusion process is simply the product of the applied learned Gaussian transitions $p_\theta (x_{t-1} | x_t)$:
\begin{equation}
    p_\theta(x_{0:T}) := p_\theta(X_T) \prod^T_{t=1}p_\theta(x_{t-1} | x_t).
\end{equation}

%%%%%%%%%%%%%%%%%%%%%%%%%%%%%%%%%%%%%%%%%%%%%%%%%%%%
% SCENESENSE DESIGN AND IMPLEMENTATION
%%%%%%%%%%%%%%%%%%%%%%%%%%%%%%%%%%%%%%%%%%%%%%%%%%%%

\section{SceneSense Design and Implementation}\label{sec:scene-sense-design-implementation}

SceneSense \cite{reed2024scenesense, reed2024online} is a diffusion model designed to generate 3D occupancy predictions given an observed occupancy map. Notably, SceneSense only generates occupancy predictions in space that has not been observed, and therefore maintains the fidelity of the observed map. In this section we will discuss the design of the SceneSense network as an unconditional diffusion model as well as our method for training. Then we will discuss our inference time control method of occupancy inpainting. 

\subsection{Network Design}
\subsubsection{Denoising Network}
We design our denoising network as an unconditional diffusion model. Specifically the denoising network is a U-net constructed from the huggingface diffusers library of blocks \cite{von-platen-etal-2022-diffusers} and consists of Resnet \cite{he2015resnet} downsampling/upsampling blocks. Notably, as discussed in \citet{reed2024scenesense, reed2024online} the addition of conditioning for guidance, as often seen in text-to-image diffusion models \cite{rombach2022high}, does not generate more accurate predictions and increases inference times in this particular application. These ablations are presented in Appendix A and B.
\begin{figure}
    \centering
    \includegraphics[width=1.0\linewidth]{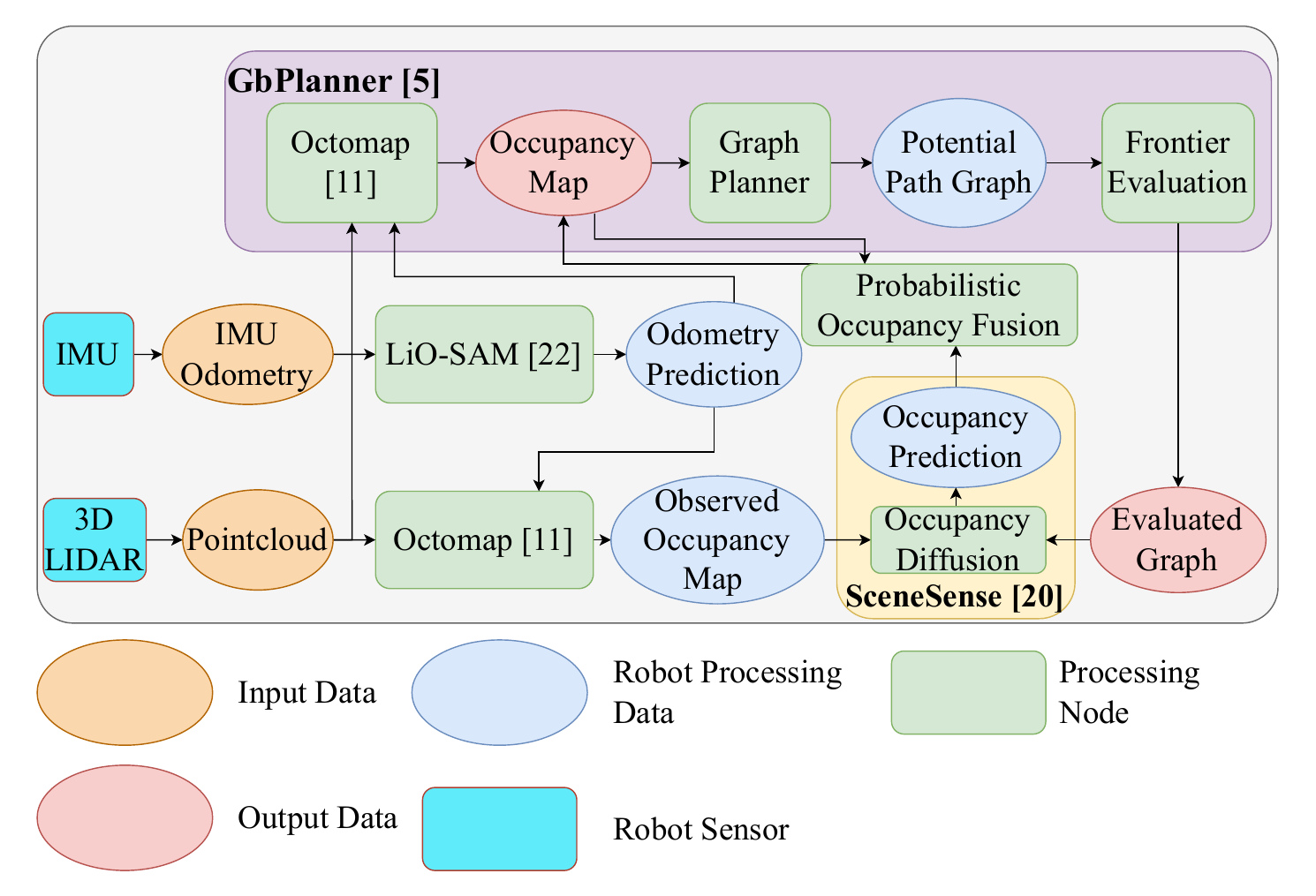}
    \caption{Block diagram showing the system design for onboard SceneSense occupancy prediction. The system is comprised of an IMU and LIDAR sensor to generate odometry and occupancy maps. Once the occupancy map is built, a graph is constructed to evaluate frontier points for occupancy prediction. Local occupancy is then subselected around these points and sent to the SceneSense framework that provides occupancy predictions. These predictions are then merged with the running occupancy map using the probabilistic update rule.}
    \label{fig:robot_archetecture}
\end{figure}

\subsubsection{Occupancy Mapping} 
Occupancy mapping allows platforms to build a running map of areas that have been measured to contain matter using onboard sensors like lidar or RGB-D cameras. For our framework we use the popular occupancy mapping framework OctoMap \cite{hornung13octomap} to generate an occupancy map as the platform explores the environment. Importantly,  OctoMap provides a probability of occupancy $o \in [0,1]$ for every voxel in the map that has been observed using pose ray casting. This means that as we explore we will maintain not only a map of occupied areas $M_o$, but also a map of areas that have been measured to not contain any data $M_u$. These maps will later be used to inform SceneSense where occupancy predictions should be made.

\subsection{Training}
During training, we generate a noisy local occupancy map $x_{t}$ where $t \in [1,T]$ from a ground truth local occupancy map $x$. We train the diffusion model $f_\theta$ to predict the noise applied to $x_t$.

\subsubsection{Data Augmentation}
In previous SceneSense implementations \cite{reed2024scenesense,reed2024online}, the primary area for prediction was centered around the robot. In this paradigm, the SceneSense training data was collected as local occupancy data collected around the robot. However, this training approach struggles to capture the variety of poses possible when doing occupancy predictions at the frontier. The graph-based planner, where poses are selected, can be much closer to structures like walls, and at a wider variety of rotation poses. To account for this, we augment the training data to more closely approximate that of the graph based poses. For each pose we generate 10 noisy pose samples where noise is added from a uniform distribution of $[-1,1]$ meters in the x,y direction and $[0, 2\pi]$ in the rotational plane. This modification increases the training set by 10$x$ but yields much more consistent results at inference time when performing occupancy prediction around frontier nodes in the graph. 

\subsubsection{Occupancy Corruption}
To corrupt each ground truth local occupancy map $x$ to train the network we add Gaussian noise to $x$ to generate $x_t$. This corruption process is defined in \ref{eq:add_noise} where the intensity of the noise is controlled by $\alpha_t$ which is configured by a linear noise scheduler \cite{ho2020denoising}. 

\subsubsection{Loss Function}
The network $f_{\theta}$ is trained using the calculated $l_2$ loss between the denoised $x_t$ prediction and the associated ground truth data $x$. $l_2$ loss is a popular diffusion loss function, however other loss functions such as cross-entropy loss or mean squared error can be applied and have had some success in similar diffusion frameworks \cite{glide-nichol22a, rombach2022high, chen2022diffusiondet, ji2023ddp}.
\label{sec:experiments}

\subsection{Inference}
\begin{figure}[]
    \centering
    \includegraphics[width=0.7\linewidth]{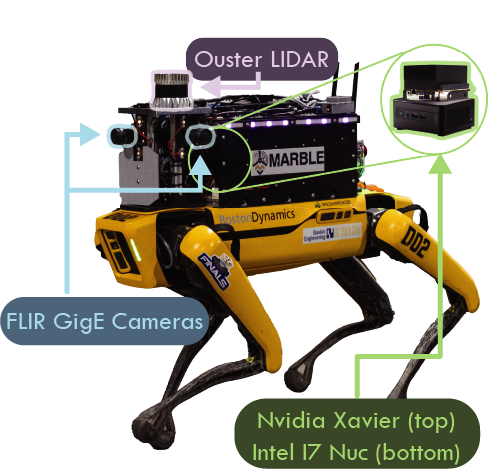}
    \caption[Spot Platform]{Spot robotic platform with onboard compute and sensors. Sensor suite consists of a 64 Beam OS1 lidar as well a 3 FLIR GigE Cameras.}
    \label{fig:spot_w_sensors}
\end{figure}
\subsubsection{Sampling Process} 
The trained noise prediction network $f_\theta$ takes isotropic Gaussian noise $\mathcal{N}(0, \mathcal{I})$ as the starting point $x_T$ to begin the reverse diffusion process. The noise is iteratively removed by using $f_\theta$ to compute $x_{t-1}$.

\subsubsection{Occupancy Inpainting}
Our method of occupancy inpainting is provided in Algorithm \ref{alg:unconditional_diffusion} which ensures observed space is never overwritten with SceneSense predictions. Inspired by image inpainting methods seen in image diffusion \cite{rombach2022high} and guided image synthesis methods \cite{meng2022sdedit}, occupancy inpainting continuously applies the known occupancy information to the diffusion target during inference. To perform occupancy inpainting we select our target region $x$ as a submap of the full occupancy map. From $x$ we build an occupied map $M_o$ and an unoccupied map $M_u$ composed of observed occupied voxels and observed unoccupied voxels, respectively. From $M_u$ and $M_o$ we generate noisy representations commensurate with the current diffusion step $t$ using Eq. \ref{eq:add_noise}, $M_{un}$ and $M_{on}$, respectively. Finally the data from $M_{un}$ and $M_{on}$ replaces the data in the diffusion target at the associated coordinates. 

This process is repeated for each inference step and can be seen in Figure \ref{fig:diffusion_framework}. This method ensures the diffusion model does not predict or modify geometry in space that has already been observed. This inpainting method is crucial for increasing the fidelity of the scene predictions.
%%%% ORIGINAL

\begin{algorithm}
\caption{Occupancy Inpainting}
\label{alg:unconditional_diffusion}
\begin{algorithmic}[1] % The [1] enables line numbering
\Require Unconditional diffusion model $\epsilon_\theta(x_t)$, $x_{obs}$, $m_{obs}$
\State $x_T \sim \mathcal{N}(0, I)$
\For{$t = T, \dots, 1$}
    \Statex \quad \textit{In-painting step:}
    \State \quad $x_{obs}^{t-1} \sim \mathcal{N}(\sqrt{\alpha_{t-1}} x_{obs}, (1 - \alpha_{t-1}) I)$
    \State \quad $x_{t-1}^{m_{obs}} := x_{obs}^{t-1}$
    \Statex \quad \textit{Standard reverse diffusion step:}
    \State \quad $\mu_\theta(x_t) := \sqrt{\frac{1}{\alpha_t}} \left( x_t - \frac{\sqrt{\beta_t}}{\sqrt{1-\alpha_t}} \epsilon_\theta(x_t) \right)$
    \State \quad $x_{t-1} \sim \mathcal{N}(\mu_\theta(x_t), \tilde{\beta}_t I)$
\EndFor
\State \textbf{return} $x_0$
\end{algorithmic}
\end{algorithm}

\subsubsection{Multiple Prediction}
\begin{figure}[t]
    \centering
    \includegraphics[width=\linewidth]{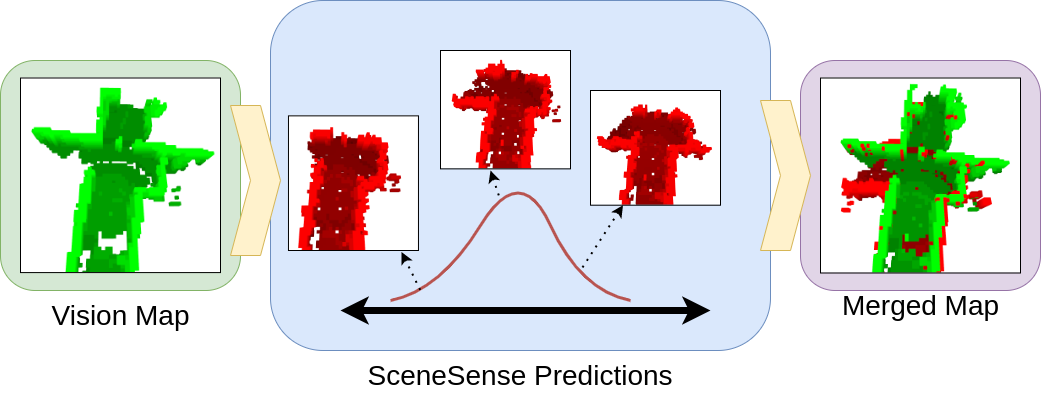}
    \caption{\textbf{Multi-Prediction Occupancy Merging}: SceneSense predicts various occupancy maps based on equivalent input data that form a distribution. This distribution forms a curve where more likely predictions occur more often, and less likely predictions occur infrequently. These predictions are fused into the merged map using Eq. \ref{eq:prob_map_merging}. The resulting merged map naturally filters out the unlikely voxel predictions, forming an extended occupancy map.}
    \label{fig:occ_voting}
\end{figure}

Diffusion is a noisy process that can generate different results given the same context. In image generation this is a desirable characteristic as the framework can generate different results given the same prompt, increasing the diversity of the generated data. As shown in Figures \ref{fig:occ_voting} and \ref{fig:Narrow Hallway Predictions}, SceneSense has the same behavior as these networks and can generate different reasonable predictions based on the same input information. Further there is no compute time increase  (assuming enough compute is available) as multiple predictions can be done in parallel. We present our method for merging these various occupancy maps in Section \ref{sec:prob_map_merging}.

%%%%%%%%%%%%%%%%%%%%%%%%%%%%%%%%%%%%%%%%%%%%%%%%%%%%
% ROBOTIC SYSTEM DESIGN
%%%%%%%%%%%%%%%%%%%%%%%%%%%%%%%%%%%%%%%%%%%%%%%%%%%%

\section{Robotic System Design}\label{sec:method_robo_sys_arch}
Our robotic system is a quadruped (Spot) as shown in Figure \ref{fig:spot_w_sensors} combined with an off-board high performance computer to handle computationally expensive requests. Much of the system design is inherited from the University of Colorado at Boulder MARBLE team that competed in the DARPA Subterranean challenge \cite{biggie2023flexible, chung2023into}. For convenience, we provide relevant design decisions in this section and a block diagram of the system in Figure \ref{fig:robot_archetecture}.

\subsection{Sensor Suite}
The equipped sensor suite was designed with the purpose of providing 3D point cloud information and sufficient data for accurate localization. The primary sensor in the Spot sensor suite is the Ouster OS1-64 LIDAR which provides 3D point clouds for mapping and localization. In addition, a LORD Microstrain 3DM-GX5-15 IMU is used to measure the linear and angular acceleration of the Ouster, for use in the localization system. 

\subsection{Localization}
Localization is required for Spot to perform volumetric mapping. We have implemented the popular LIDAR-based localization method LIO-SAM \cite{shan2020liosam} to provide localization at runtime. To improve localization performance both the IMU and lidar sensor were fastened to a 6061 aluminum sensor plate. This allowed for a high-precision relative transform between the sensors, reducing the necessity of extrinsic calibration. Furthermore, LIO-SAM requires aligned sensor timestamps and sensor publish rates to be constant. The LORD IMU however allows for large fluctuation in publish rate due to the USB transmission delays. To reduce the sensitivity the IMU timestamps are adjusted when messages are not received within $15\%$ of the nominal rate. Additionally, the lidar sensor uses Precision Time Protocol (PTP) to synchronize with the onboard computer. These modifications allowed for consistent localization using LIO-SAM. 
\subsection{Frontier Identification and Evaluation}
The first work in this domain \cite{reed2024scenesense} focused on occupancy prediction around the robot. Later work \cite{reed2024online} updates the occupancy prediction framework to predict occupancy at any point in the map, demonstrating enhanced utility for predictions at range. We leverage this enhancement, identifying interesting areas for prediction using the graph-based exploration planner GBPlanner \cite{dang2020graph}. GBPlanner builds a graph where nodes are potential exploration points and edges are feasible paths to navigate from node to node. A ray casting algorithm is run at each node in the graph to quantify the number of observed, free, and unknown voxels in that node's field of view. After finding the shortest path to each node the \emph{Exploration Gain} can be calculated for each node in the graph as follows: 
\begin{multline}
\label{eq:GB_planner}
    \textbf{ExplorationGain}(\sigma_i) = e^{-\gamma_\mathcal{S}\mathcal{S}(\sigma_i,\sigma_{exp})}  \\
    \cdot \sum_{j = 1}^{m_i}\textbf{VolumetricGain}(v^i_j)e^{-\gamma_\mathcal{D}\mathcal{D}(v^i_1,v^i_j)},\end{multline}
where $\mathcal{S}(\sigma_i, \sigma_{exp})$, $\mathcal{D}(v^i_1,v^i_j)$ are weight functions with tunable factors $\gamma_\mathcal{S}, \gamma_\mathcal{D} > 0$ respectively. Furthermore $\mathcal{D}(v^i_1,v^i_j)$ is the cumulative Euclidean distance from a vertex $v^i_j$ to the root $v^i_1$ along a path $\gamma_i$.

We use exploration gain to rank nodes where occupancy prediction should run. Given a minimum node spacing $d_{m}$ and a maximum number of frontier prediction nodes $n_{max}$, SceneSense generates occupancy predictions at range, centered around the identified frontiers.

\subsection{Mapping}
We use the probabilistic volumetric mapping method OctoMap \cite{hornung13octomap} as the system's mapping framework. OctoMap was adopted for its log-odds update method for predicting occupied and unoccupied cells. This approach allows for computationally efficient fusion of observed occupancy and predicted occupancy maps. Further discussion on map fusion is provided in Section \ref{sec:prob_map_merging}.

%%%%%%%%%%%%%%%%%%%%%%%%%%%%%%%%%%%%%%%%%%%%%%%%%%%%%%%%%%%
% PROBABILISTIC MAP MERGING
%%%%%%%%%%%%%%%%%%%%%%%%%%%%%%%%%%%%%%%%%%%%%%%%%%%%%%%%%%%

% \begin{figure}
%     \centering
%     \begin{subfigure}{0.4\textwidth}
%         \centering
%         \resizebox{0.8\linewidth}{!}{\input{figures/floor_8_iou_2.tikz}}
%         \caption{}
%         \label{fig:subfig1_startup}
%     \end{subfigure}
%     \hfill
%     \begin{subfigure}{0.4\textwidth}
%         \centering
%         \resizebox{0.8\linewidth}{!}{\input{figures/floor_8_iou_1.tikz}}
%         \caption{}
%         \label{fig:subfig2_Startup}
%     \end{subfigure}
%     \caption{Env 2 IoU Probability Mass Function (PMF). (a) IoU histogram of RC SceneSense predictions. (b) IoU histogram of FC SceneSense predictions. The IoU distributions show that RC occupancy predictions are more likely to be similar than FC predictions.}
%     \label{fig:pred_PMF}
% \end{figure}

\section{Probabilistic Map Merging} \label{sec:prob_map_merging}
In previous work, predicted occupancy was merged into the running map in a “fire and forget” fashion \cite{reed2024scenesense}. Each prediction was incorporated by setting its occupied cells to 1, but when a new prediction was generated, the previous one was discarded and replaced. While simple, this approach prevents the system from maintaining a coherent and continuous estimate of the environment. A single out of distribution prediction can temporarily warp the map, causing the planner to divert the robot along a spurious path. Once the next prediction arrives and overwrites the first, the planner abruptly redirects, potentially producing oscillatory shifts in the planned trajectory. To address these issues, we modify the probabilistic occupancy update formula presented for the OctoMap mapping framework \cite{hornung13octomap}.

We define the probability that a voxel $m$ is occupied given an occupancy prediction $d_t$ or sensor reading $z_t$ as $P(m|d_t)$ or $P(m|z_t)$, respectively. The set of sensor estimates $z_{1:t}$ and diffusion estimates $d_{1:t}$ populate the joint set $\{z_{1:t}, d_{1:t}\}$ which we denote as  $j_{1:t}$. As discussed in \cite{reed2024scenesense}, SceneSense only operates on voxels $m$ that are not contained in the observed set $\mathcal{O}$, where $z_{t:t-1} = \varnothing$, and therefore $P(m|j_{1:t})$ will never require an update given $P(m|d_t)$ and $P(m|z_t)$ at the same time. As such we generate the piece-wise update rule for merging diffusion into the running occupancy map.

\begin{equation} \label{eq:prob_map_merging}
\begin{aligned}
P(m \mid j_{1:t})= \\
& \hspace{-60pt} \begin{cases} 
\left[ 1 + \frac{1-P(m \mid d_t)}{P(m \mid d_t)}\frac{1-P(m \mid j_{1:t-1})}{P(m \mid j_{1:t-1})} \frac{P(m)}{1-P(m)}\right]^{-1} & \text{if } m \notin \mathcal{O}  \\
\left[ 1 + \frac{1-P(m \mid z_t)}{P(m \mid z_t)}\frac{1-P(m \mid j_{1:t-1})}{P(m \mid j_{1:t-1})} \frac{P(m)}{1-P(m)}\right]^{-1} & \text{if } m \in \mathcal{O}.
\end{cases}
\end{aligned}
\end{equation}

\smallskip
In this framework $P(m|z_t)$ and $P(m|d_t)$ can be configured to different values prior to runtime. Generally, $P(m|d_t)$ given a predicted occupied cell is set lower than $P(m|z_t)$ given a sensed occupied cell, as we trust the sensor more than our generative model. By using this probabilistic approach to map merging, the final merged map benefits from prediction persistence as the system explores as well as increased map fidelity due to multi-prediction occupancy refinement.  

%%%%%%%%%%%%%%%%%%%%%%%%%%%%%%%%%%%%%%%%%%%%%%%%%%%%%%%%
% OCCUPANCY PREDICTION EXPERIMENTS AND EVALUATIONS
%%%%%%%%%%%%%%%%%%%%%%%%%%%%%%%%%%%%%%%%%%%%%%%%%%%%%%%%

\section{Occupancy Prediction Experiments and Evaluations}\label{sec:occ_pred_experiments_evals}
In this section, we provide results and evaluations of the modified SceneSense occupancy prediction framework onboard a real-world system. In particular we evaluate if the SceneSense enhanced occupancy maps better represent a set of fully observed ground truth maps.

\vspace{5pt}
\textbf{Training and Implementation}.
To train SceneSense, we collected real-world occupancy maps from various indoor environments across different buildings. We gathered approximately 1 hour worth of occupancy data, resulting in $11,296$ unique poses with associated complete local occupancy maps. Any areas that were used to train the model are omitted from the results presented here. 

We implement the same denoising network structure presented in \citet{reed2024scenesense}. It is a U-net constructed from the HuggingFace Diffusers library of blocks \cite{von-platen-etal-2022-diffusers} and consists of Resnet \cite{he2015resnet} downsampling/upsampling blocks. The diffusion model is trained using randomly shuffled pairs of ground truth local occupancy maps $x$. We use Chameleon cloud computing resources \cite{keahey2020chameleon} to train our model on one A100 with a batch size of $32$ for $250$ epochs or $88,208$ training steps. We use a cosine learning rate scheduler with a 500 step warm up from $10^{-6}$ to $10^{-4}$. The noise scheduler for diffusion is set to $1000$ noise steps. 

At inference time we evaluate our dataset using an RTX 4070 TI Super GPU for acceleration. The number of diffusion steps is configured to $30$ steps.

\vspace{5pt}
\textbf{Evaluation Metrics}.
We evaluate predicted occupancy using a classical computer vision metric, Intersection over Union (IoU) \cite{everingham2010pascal}, and newer generative modeling metrics, Fréchet Inception Distance (FID) \cite{heusel2017gansFID} and Kernel Inception Distance (KID) \cite{binkowski2018demystifyingKID}. IoU provides a direct measure of overlap between predicted and ground truth occupied voxels. To assess the realism and diversity of generated occupancy predictions, we report FID and KID. FID measures the Fréchet distance between Gaussian statistics of features extracted from predicted and real samples, while KID uses a polynomial kernel Maximum Mean Discrepancy, rather than assuming Gaussian distributions, to provide an unbiased estimate. Both metrics have become standard in generative modeling, and we compute them using the \texttt{clean-fid} library \cite{parmar2021cleanfid}.

It is worth noting that generating good metrics to evaluate generative frameworks remains a difficult task \cite{naeem2020reliable}. FID and KID have become standard metrics for many generative methods due to their ability to score accuracy of predicted results and diversity or coverage of the results when compared to a set of ground truth data. While these metrics are fairly new to robotics, which traditionally evaluates occupancy data with metrics like accuracy, precision and IoU, we report FID and KID, in addition to IoU, because they have been shown to be an effective measure for evaluating predicted scenes \cite{reed2024scenesense, tang2023diffuscene}.

\subsection{SceneSense Generative Occupancy Evaluation}
\begin{figure}[]
    \centering
    \begin{subfigure}{0.45\linewidth}
        \centering
        \includegraphics[width=\linewidth, trim={50pt 0 5cm 0}, clip]{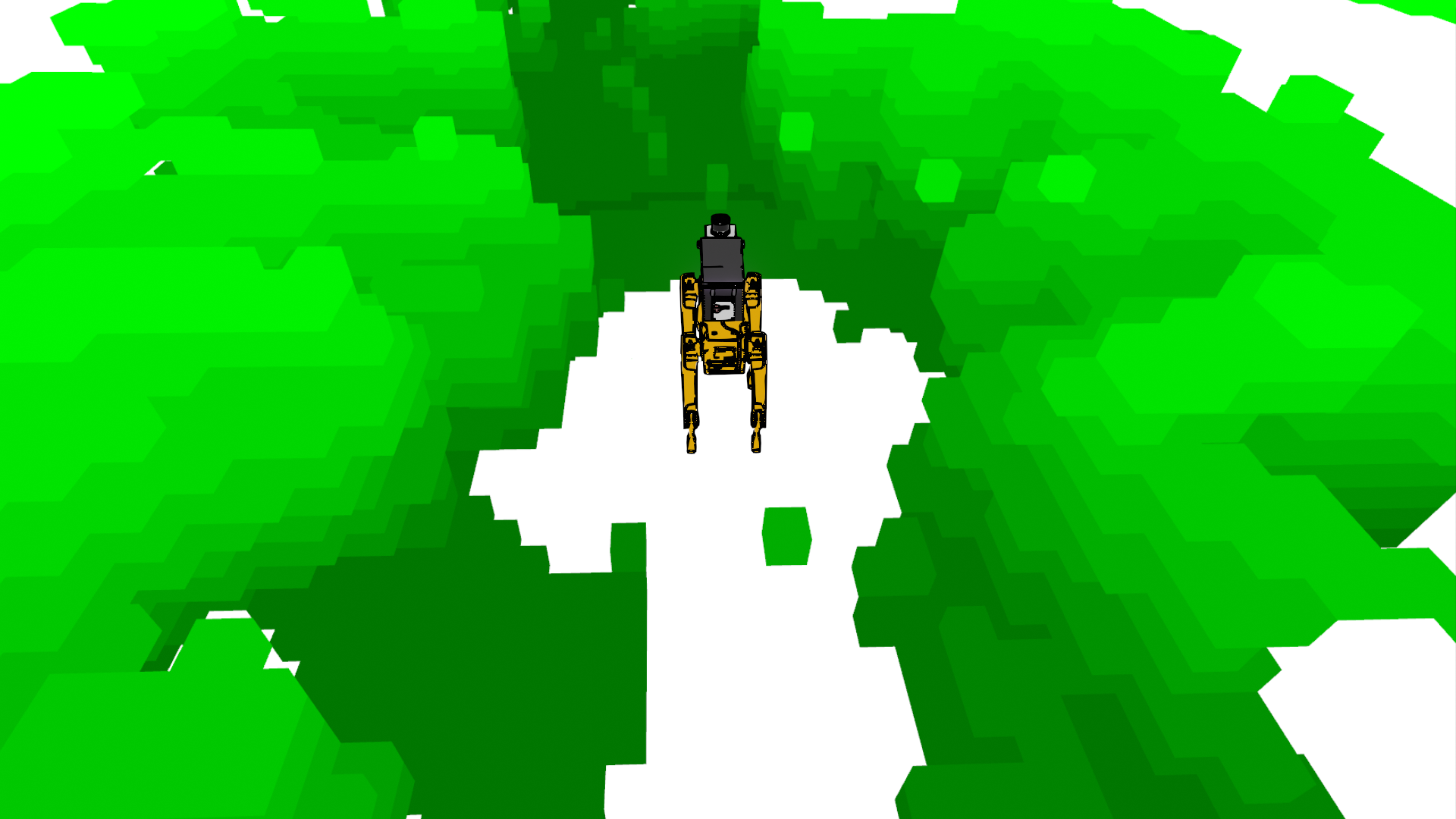}
        \caption{}
        \label{fig:subfig1_startup}
    \end{subfigure}
    \hfill
    \begin{subfigure}{0.45\linewidth}
        \centering
        \includegraphics[width=\linewidth, trim={50pt 0 50pt 0}, clip]{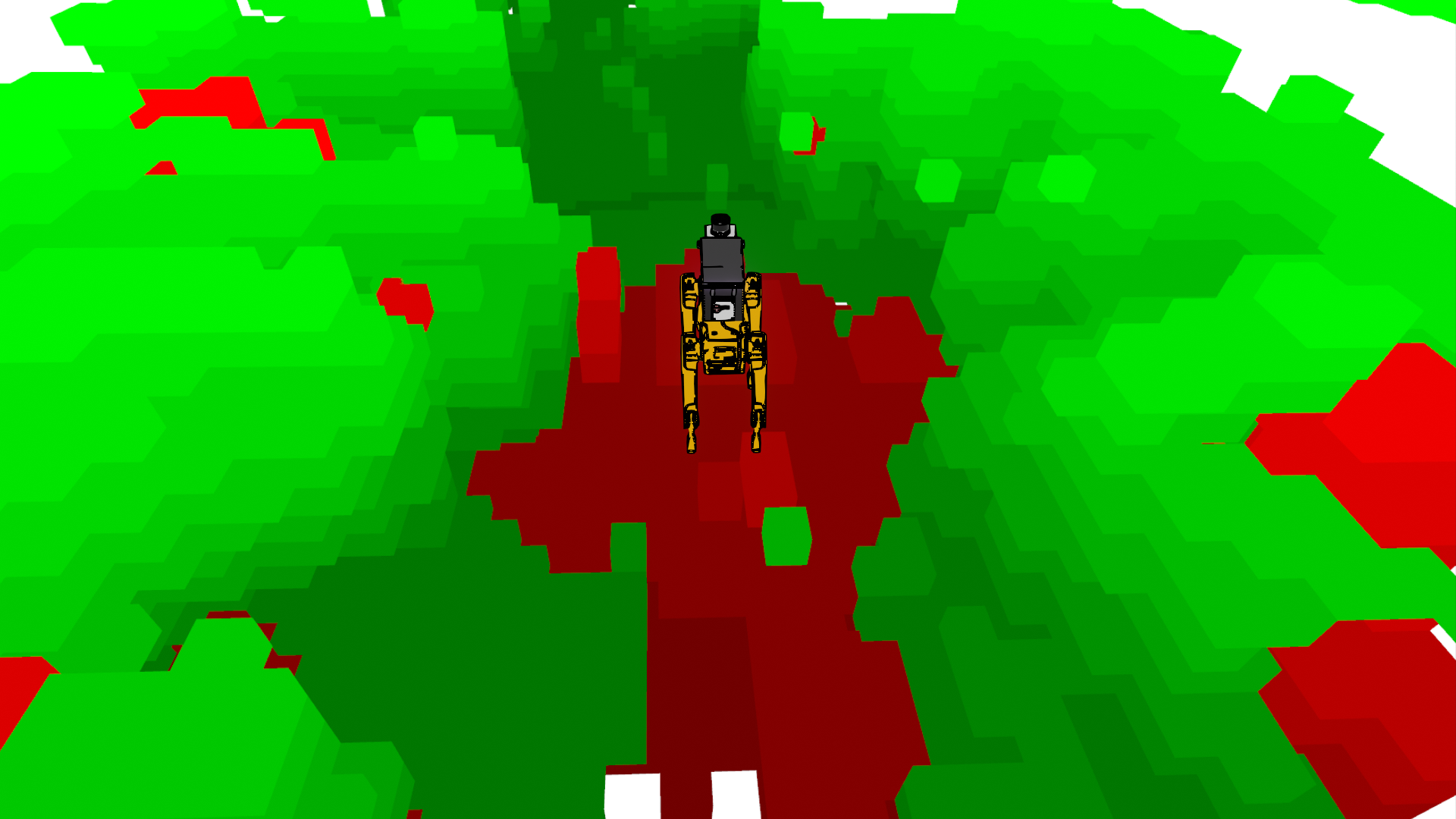}
        \caption{}
        \label{fig:subfig2_startup}
    \end{subfigure}
    \caption{At startup, the robot cannot observe the ground directly under it due to the mounting location of the lidar. (b) SceneSense generates occupancy predictions (Red Voxels) that fill in the hole under the robot as well as some of the vertical occluded geometry. With the additional predictions, the robot autonomously generates a traversable path and begins exploring without the need for manual intervention via teleoperation.}
    \label{fig:startup-figure}
\end{figure}

\begin{figure}[]
    \centering
    \begin{subfigure}{0.45\linewidth}
        \centering
        \includegraphics[width=\linewidth]{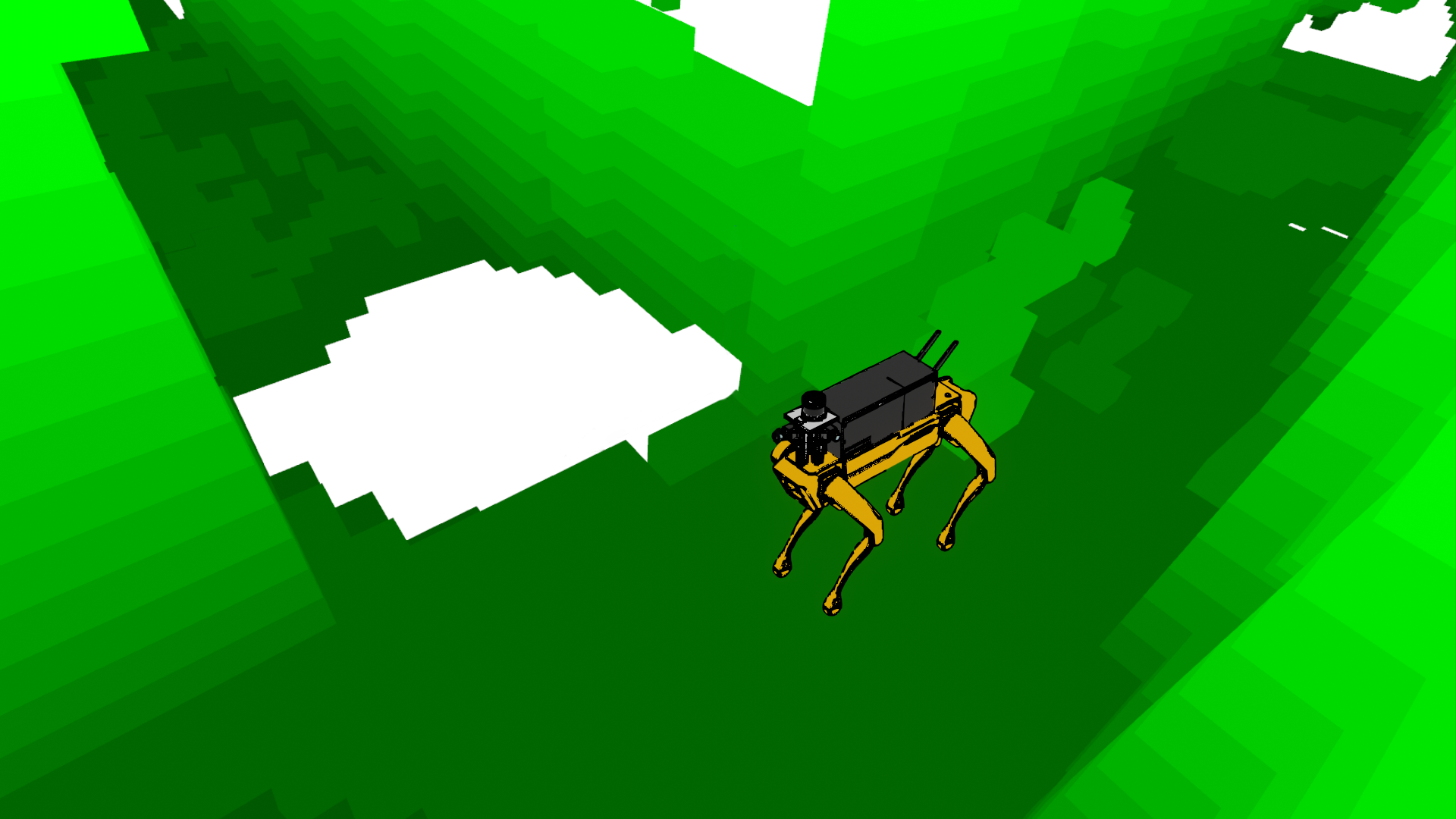}
        \caption{}
        \label{fig:hallway_missed}
    \end{subfigure}
    \hfill
    \begin{subfigure}{0.45\linewidth}
        \centering
        \includegraphics[width=\linewidth]{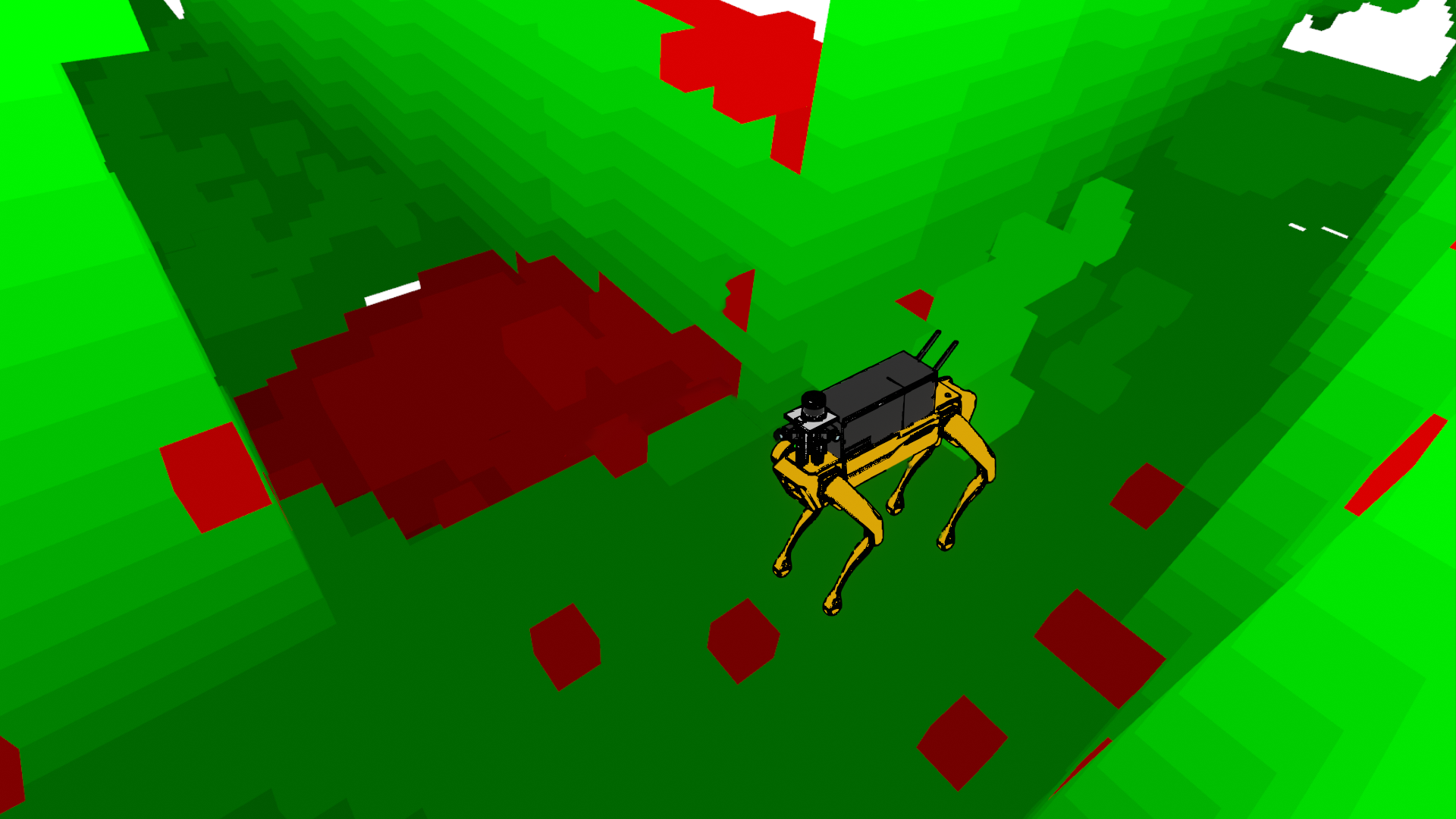}
        \caption{}
        \label{fig:subfig2_hall}
    \end{subfigure}
    \caption{(a) During traversal, the robot fails to observe the ground directly to its right as a result of the mounting location of the lidar and is unable to plan down the hallway for exploration. (b) SceneSense generates occupancy predictions (Red Voxels) that fill in the hole to the right of the robot as well as some of the vertically occluded geometry. With the additional predictions, a plan can now be generated, allowing the robot to continue navigation down the hallway.}
    \label{fig:hallway_traversal}
\end{figure}

\begin{figure}[]
    \centering
    \begin{subfigure}{0.3\linewidth}
        \centering
        \includegraphics[width=\linewidth]{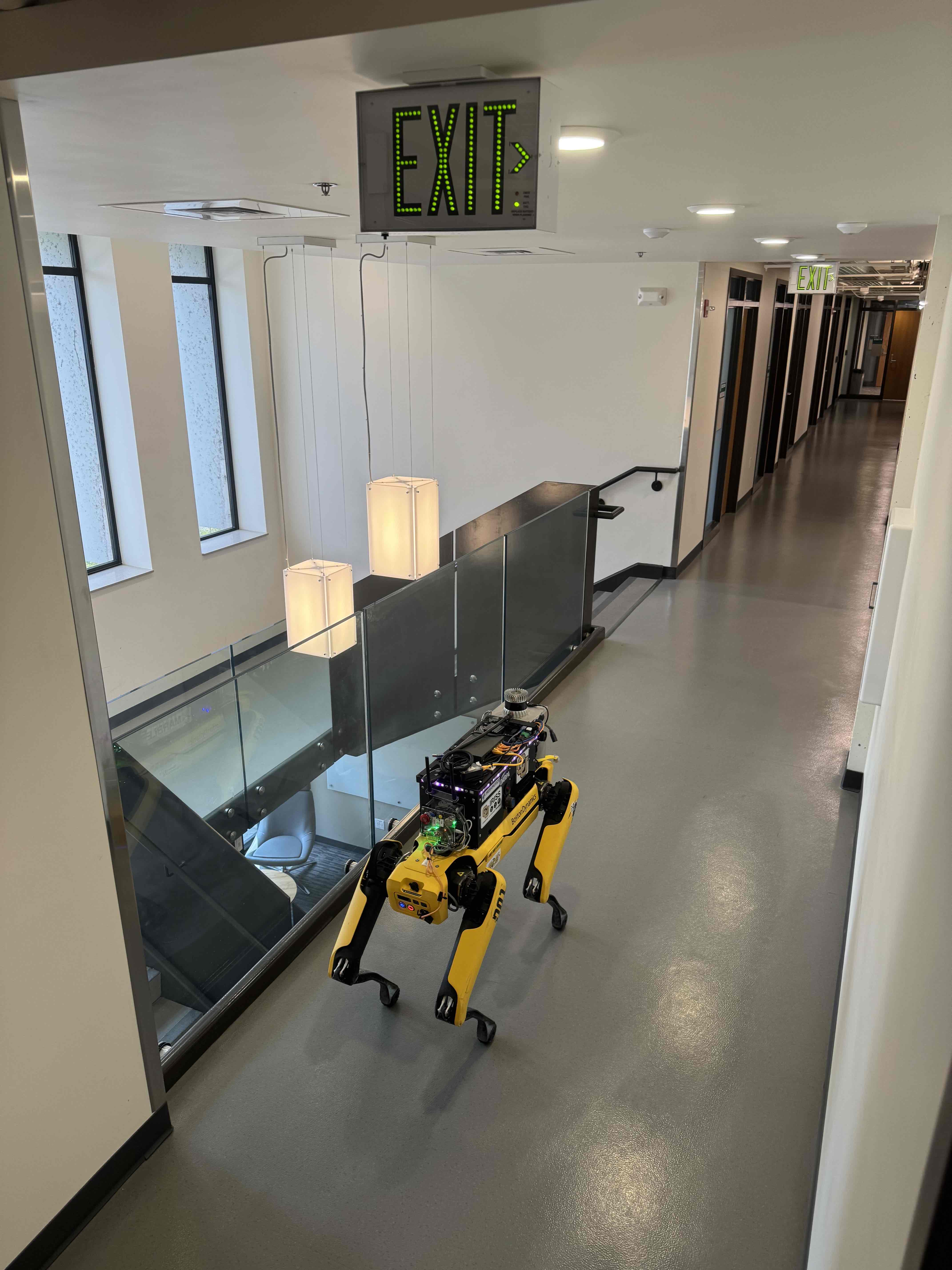}
        \caption{}
        \label{fig:stairs_pic}
    \end{subfigure}
    \hfill
    \begin{subfigure}{0.6\linewidth}
        \centering
        \includegraphics[width=\linewidth]{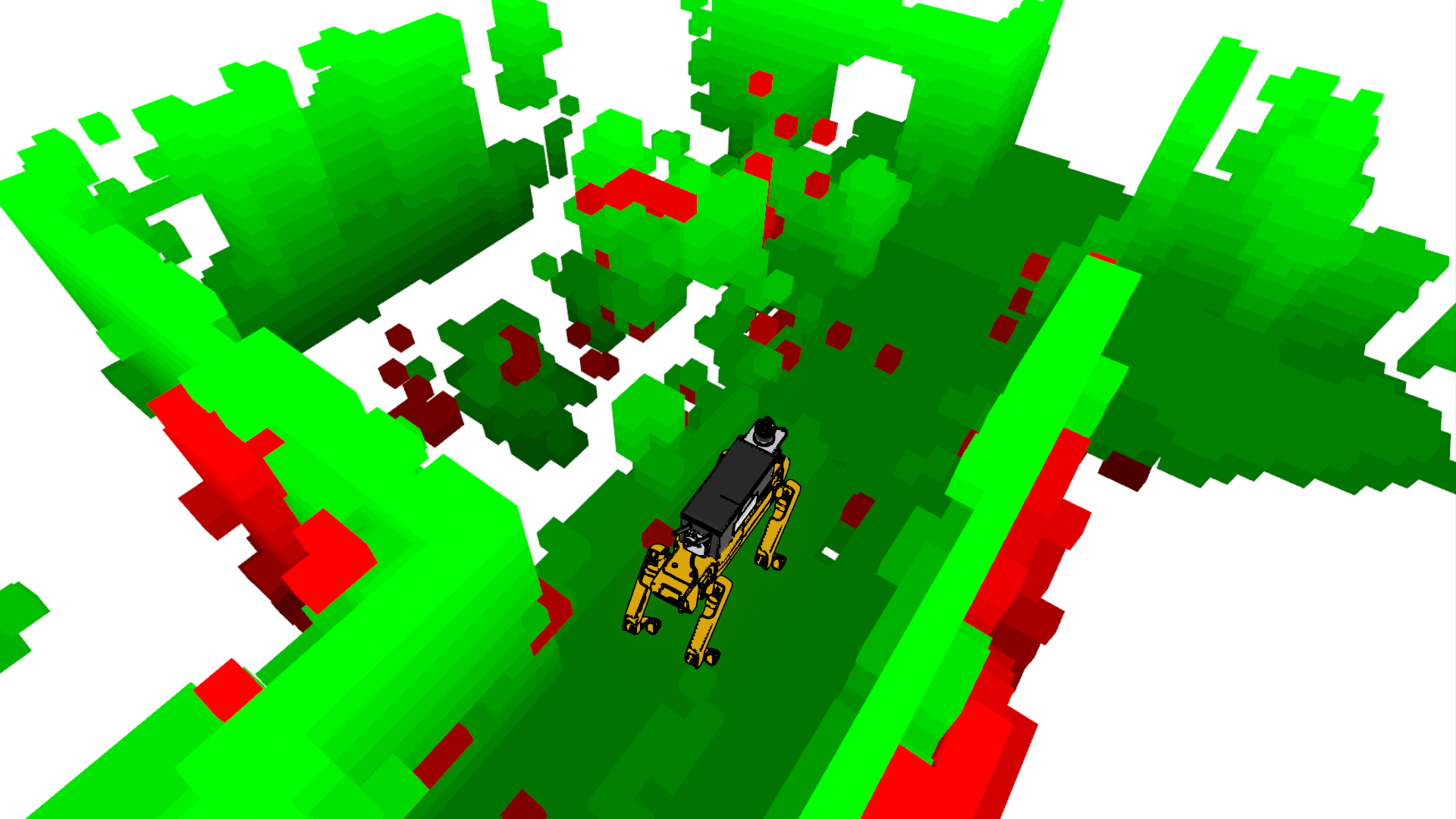}
        \caption{}
        \label{fig:stairs_diff}
    \end{subfigure}
    \caption{(a) The robot's position during exploration near a glass railing, an adversarial scenario for planners that do not consider ground traversability. (b) The running occupancy map shows that the ground suddenly ends to the right of the robot, where the railing prevents it from navigating off the edge. Samplers that do not check for traversability—effective in solving scenarios in Figures \ref{fig:hallway_traversal} and \ref{fig:startup-figure}—could mistakenly allow paths through open space. Here, SceneSense correctly generates red voxels to indicate untraversable ground, preventing unsafe navigation.}
    \label{fig:stairs_diff_all}
\end{figure}

For the following experiments we evaluate the occupancy generation capabilities of SceneSense onboard a real-world robot in two unique test environments. In particular, we examine the fidelity of predictions around the robot with predictions at the frontiers of the map, ablating the map update methods and the running sensor only map.
\vspace{5pt}\\
\textbf{Experimental Setup}. \ \ SceneSense predictions are evaluated in 2 environments. Environment 1 is a long hallway with cutouts for classrooms and 1 right turn. Select frames shown in Figures \ref{fig:startup-figure} and \ref{fig:hallway_traversal} are from Environment 1. Environment 2 consists of similar carpeted area with 4 hallways and 4 turns, forming a square shape. We evaluate the occupancy prediction framework using the following test configurations. 
\begin{enumerate}
    \item \textbf{Baseline or SceneSense}: A comparison between Octomap sensor-only local occupancy (BL) with the SceneSense occupancy prediction included (SS). 
    \item \textbf{Robot-centric or Frontier-centric}: Robot-centric diffusion (RC) predicts occupancy at a radius of $3.3m$ about the robot while frontier-centric diffusion (FC) predicts occupancy at a radius of $3.3m$ at an identified location in the map, which has a maximum range of 7$m$ from the robot. 
    \item \textbf{One Shot Map Merging or Probabilistic Map Merging}: One shot map merging (OSMM) simply takes the current local occupancy map and a SceneSense occupancy prediction and populates the predicted occupancy information in the running map using the ``fire and forget" approach described in Section \ref{sec:prob_map_merging}. Probabilistic map merging (PMM) keeps a running local merged occupancy map that uses update Equation \ref{eq:prob_map_merging} to update the occupancy map for every new occupancy prediction. In practice, each pose will receive 3-5 SceneSense predictions to merge into the running map.
\end{enumerate}

\begin{figure}[]
    \centering
    % First subfigure (full-width map)
    \begin{subfigure}{0.8\linewidth}
        \centering
        \includegraphics[width=\linewidth]{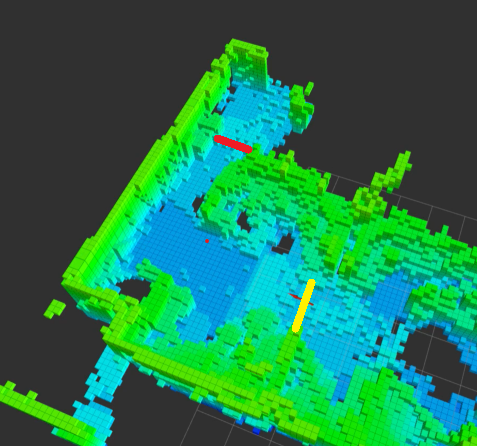}
        \caption{}
        \label{subfig:lab_map}
    \end{subfigure}
    
    \vspace{1em} % Adds some vertical space between subfigures

    % Second subfigure (side-by-side images)
    \begin{subfigure}{0.9\linewidth}
        \centering
        \includegraphics[width=0.4\linewidth, trim={0 0 0 40pt}, clip]{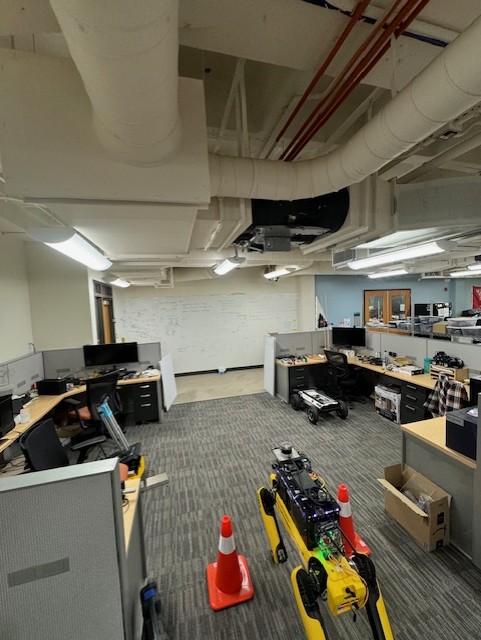}
        \hfill
        \includegraphics[width=0.4\linewidth, trim={0 20pt 0 20pt}, clip]{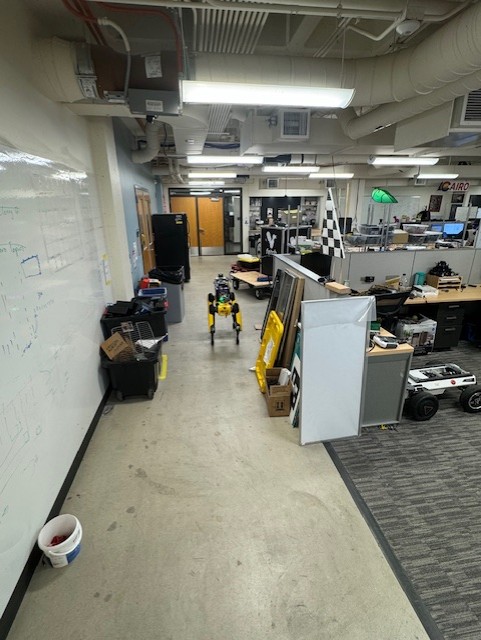}
        \caption{}
        \label{subfig:spot_pictures}
    \end{subfigure}

    \caption{\textbf{Test Environment}: The test environment used to produce Table \ref{table:hallway_traversal_results}. (a) The complete occupancy map representation with the start and finish line shown in yellow and red, respectively. (b) The starting position (left) and finish position (right) of the Spot robot.}
    \label{fig:narrow-corner_Test}
\end{figure}

\textbf{Generative Occupancy Results Discussion}. The results in Table \ref{table:sim_results} show that RC predictions are quite similar between OSMM and PMM approaches, reducing the FID of the environments by an average of $28.5\%$ and $25\%$ for OSMM and PMM, respectively. These results are similar to the simulation-based results presented in \cite{reed2024scenesense}. However, the FC results show that PMM achieves a much greater improvement than OSMM. This improvement is reflected in the average FID, where a lower value indicates better performance. PMM reduces the average FID by $75\%$, whereas OSMM achieves only an $11\%$ reduction.

\setlength{\tabcolsep}{3pt} 
\begin{table}[h]
\footnotesize % Ensures the table fits in a single column
\centering
\caption{\textbf{Comparison of Running Occupancy (BL) vs. SceneSense (SS)}. Evaluations of each method are provided as robot-centric (RC) and frontier-centric (FC) generations.}
\label{table:sim_results}

\begin{tabular}{l | c c | c c } 
\toprule
\textbf{Method} & \multicolumn{2}{c|}{\textbf{Env. 1}} & \multicolumn{2}{c}{\textbf{Env. 2}} \\
& \textbf{FID} $\downarrow$ & \textbf{KID}$\times$1000 $\downarrow$ & \textbf{FID} $\downarrow$ & \textbf{KID}$\times$1000 $\downarrow$ \\
\midrule
BL-RC          & 36.0  & 16.4  & 30.3  & 16.3  \\ 
SS-RC-OSMM     & \textbf{26.3} & \textbf{7.7}  & \textbf{20.8} & 10.1  \\
SS-RC-PMM      & 29.2  & 10.4  & 21.0  & \textbf{9.1} \\
\midrule
BL-FC          & 116.9 & 81.6  & 150.6 & 118.8 \\ 
SS-FC-OSMM     & 104.2 & 66.3  & 133.4 & 104.4 \\
SS-FC-PMM      & \textbf{30.1}  & \textbf{10.3} & \textbf{34.5} & \textbf{9.0} \\
\bottomrule
\end{tabular}
\end{table}

Interestingly, The KID values are nearly identical between SS-RC-PMM and SS-FC-PMM, indicating that the model occupancy predictions at range are as reasonable as the predictions made around the robot, even though there is less information for the predictions at range. KID is known to be less sensitive to outliers when compared to FID  \cite{bińkowski2018demystifying}. The unreasonable predictions that can occur when performing FC predictions are likely better filtered out by the KID metric, resulting in similar scores.

\begin{figure*}
    \centering
    \includegraphics[width=\linewidth]{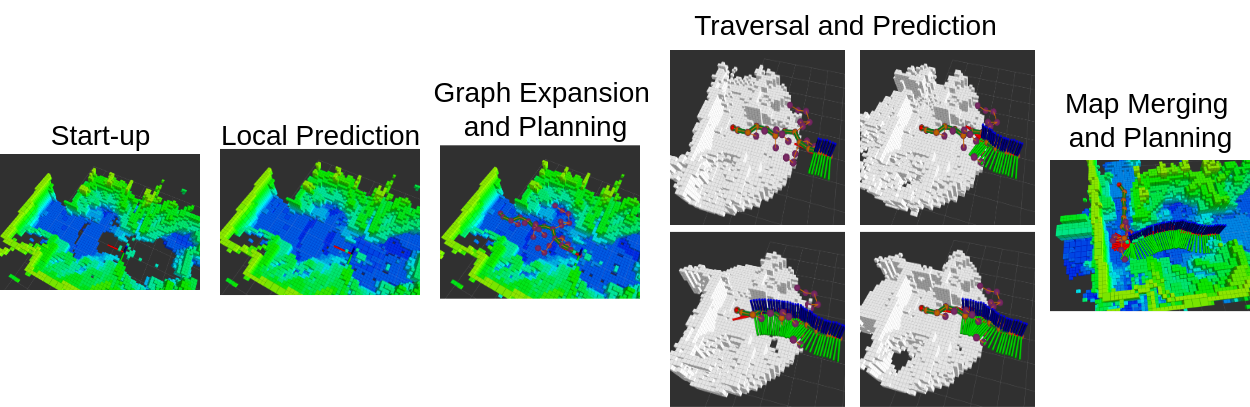}
    \caption[Frontier Planning Process]{\textbf{Example traversal of GBPlanner + Frontier SS}. At startup local diffusion is run to allow for immediate planning. Once a graph is constructed and a path is planned the robot begins both traversing the path and making predictions of the frontier occupancy (where predictions are shown in grey). Upon reaching its destination, the graph is extended based on the observed and predicted occupancy and the robot can continue exploring. }
    \label{fig:Narrow Hallway Predictions}
\end{figure*}
The large discrepancy between OSMM and PMM results when evaluated at frontiers is due to the sparsity of the occupancy map at the frontier. We predict that as the number of unknown voxels grow, so too does the distribution of predicted scenes. Intuitively, if there are no observed voxels to guide the prediction SceneSense will predict a wide variety of possible occupancy maps. Conversely, if all voxels in the target space are observed, the same occupancy map will be generated every time.

\begin{figure}
    \centering
    \begin{subfigure}{0.4\textwidth}
        \centering
        \resizebox{0.8\linewidth}{!}{% This file was created with tikzplotlib v0.10.1.
\begin{tikzpicture}

\definecolor{darkgray176}{RGB}{176,176,176}
\definecolor{steelblue31119180}{RGB}{31,119,180}

\begin{axis}[
tick align=outside,
tick pos=left,
x grid style={darkgray176},
xlabel={IoU},
xmin=0, xmax=0.6,
xtick style={color=black},
y grid style={darkgray176},
ylabel={Probability},
ymin=0, ymax=0.35,
ytick style={color=black}
]
\draw[draw=black,fill=steelblue31119180,fill opacity=0.75] (axis cs:0.0887610437641257,0) rectangle (axis cs:0.126723146674795,0.00879120879120879);
\draw[draw=black,fill=steelblue31119180,fill opacity=0.75] (axis cs:0.126723146674795,0) rectangle (axis cs:0.164685249585464,0.0485958485958486);
\draw[draw=black,fill=steelblue31119180,fill opacity=0.75] (axis cs:0.164685249585464,0) rectangle (axis cs:0.202647352496133,0.14017094017094);
\draw[draw=black,fill=steelblue31119180,fill opacity=0.75] (axis cs:0.202647352496133,0) rectangle (axis cs:0.240609455406803,0.271306471306471);
\draw[draw=black,fill=steelblue31119180,fill opacity=0.75] (axis cs:0.240609455406803,0) rectangle (axis cs:0.278571558317472,0.264713064713065);
\draw[draw=black,fill=steelblue31119180,fill opacity=0.75] (axis cs:0.278571558317472,0) rectangle (axis cs:0.316533661228141,0.155799755799756);
\draw[draw=black,fill=steelblue31119180,fill opacity=0.75] (axis cs:0.316533661228141,0) rectangle (axis cs:0.35449576413881,0.0581196581196581);
\draw[draw=black,fill=steelblue31119180,fill opacity=0.75] (axis cs:0.35449576413881,0) rectangle (axis cs:0.39245786704948,0.0146520146520147);
\draw[draw=black,fill=steelblue31119180,fill opacity=0.75] (axis cs:0.39245786704948,0) rectangle (axis cs:0.430419969960149,0.00537240537240537);
\draw[draw=black,fill=steelblue31119180,fill opacity=0.75] (axis cs:0.430419969960149,0) rectangle (axis cs:0.468382072870818,0.00366300366300366);
\draw[draw=black,fill=steelblue31119180,fill opacity=0.75] (axis cs:0.468382072870818,0) rectangle (axis cs:0.506344175781487,0.00854700854700855);
\draw[draw=black,fill=steelblue31119180,fill opacity=0.75] (axis cs:0.506344175781487,0) rectangle (axis cs:0.544306278692157,0.00488400488400488);
\draw[draw=black,fill=steelblue31119180,fill opacity=0.75] (axis cs:0.544306278692157,0) rectangle (axis cs:0.582268381602826,0.00390720390720391);
\draw[draw=black,fill=steelblue31119180,fill opacity=0.75] (axis cs:0.582268381602826,0) rectangle (axis cs:0.620230484513495,0.00317460317460317);
\draw[draw=black,fill=steelblue31119180,fill opacity=0.75] (axis cs:0.620230484513495,0) rectangle (axis cs:0.658192587424164,0.00268620268620269);
\draw[draw=black,fill=steelblue31119180,fill opacity=0.75] (axis cs:0.658192587424165,0) rectangle (axis cs:0.696154690334834,0.00146520146520147);
\draw[draw=black,fill=steelblue31119180,fill opacity=0.75] (axis cs:0.696154690334834,0) rectangle (axis cs:0.734116793245503,0.00244200244200244);
\draw[draw=black,fill=steelblue31119180,fill opacity=0.75] (axis cs:0.734116793245503,0) rectangle (axis cs:0.772078896156172,0.000732600732600733);
\draw[draw=black,fill=steelblue31119180,fill opacity=0.75] (axis cs:0.772078896156172,0) rectangle (axis cs:0.810040999066841,0);
\draw[draw=black,fill=steelblue31119180,fill opacity=0.75] (axis cs:0.810040999066841,0) rectangle (axis cs:0.848003101977511,0.000976800976800977);
\end{axis}

\end{tikzpicture}}
        \caption{}
        \label{fig:subfig1_startup}
    \end{subfigure}
    \hfill
    \begin{subfigure}{0.4\textwidth}
        \centering
        \resizebox{0.8\linewidth}{!}{% This file was created with tikzplotlib v0.10.1.
\begin{tikzpicture}

\definecolor{darkgray176}{RGB}{176,176,176}
\definecolor{steelblue31119180}{RGB}{31,119,180}

\begin{axis}[
tick align=outside,
tick pos=left,
x grid style={darkgray176},
xlabel={IoU},
xmin=0, xmax=0.6,
xtick style={color=black},
y grid style={darkgray176},
ylabel={Probability},
ymin=0, ymax=0.35,
ytick style={color=black}
]
\draw[draw=black,fill=steelblue31119180,fill opacity=0.75] (axis cs:0.00477232054086299,0) rectangle (axis cs:0.0545337045138199,0.289768483943241);
\draw[draw=black,fill=steelblue31119180,fill opacity=0.75] (axis cs:0.0545337045138199,0) rectangle (axis cs:0.104295088486777,0.301157580283794);
\draw[draw=black,fill=steelblue31119180,fill opacity=0.75] (axis cs:0.104295088486777,0) rectangle (axis cs:0.154056472459734,0.203883495145631);
\draw[draw=black,fill=steelblue31119180,fill opacity=0.75] (axis cs:0.154056472459734,0) rectangle (axis cs:0.20381785643269,0.0974607916355489);
\draw[draw=black,fill=steelblue31119180,fill opacity=0.75] (axis cs:0.20381785643269,0) rectangle (axis cs:0.253579240405647,0.0287528005974608);
\draw[draw=black,fill=steelblue31119180,fill opacity=0.75] (axis cs:0.253579240405647,0) rectangle (axis cs:0.303340624378604,0.0168035847647498);
\draw[draw=black,fill=steelblue31119180,fill opacity=0.75] (axis cs:0.303340624378604,0) rectangle (axis cs:0.353102008351561,0.00317401045556385);
\draw[draw=black,fill=steelblue31119180,fill opacity=0.75] (axis cs:0.353102008351561,0) rectangle (axis cs:0.402863392324518,0.00261389096340553);
\draw[draw=black,fill=steelblue31119180,fill opacity=0.75] (axis cs:0.402863392324518,0) rectangle (axis cs:0.452624776297475,0.000933532486930545);
\draw[draw=black,fill=steelblue31119180,fill opacity=0.75] (axis cs:0.452624776297475,0) rectangle (axis cs:0.502386160270432,0.0020537714712472);
\draw[draw=black,fill=steelblue31119180,fill opacity=0.75] (axis cs:0.502386160270432,0) rectangle (axis cs:0.552147544243388,0.00298730395817774);
\draw[draw=black,fill=steelblue31119180,fill opacity=0.75] (axis cs:0.552147544243388,0) rectangle (axis cs:0.601908928216345,0.00186706497386109);
\draw[draw=black,fill=steelblue31119180,fill opacity=0.75] (axis cs:0.601908928216345,0) rectangle (axis cs:0.651670312189302,0.000373412994772218);
\draw[draw=black,fill=steelblue31119180,fill opacity=0.75] (axis cs:0.651670312189302,0) rectangle (axis cs:0.701431696162259,0.00168035847647498);
\draw[draw=black,fill=steelblue31119180,fill opacity=0.75] (axis cs:0.701431696162259,0) rectangle (axis cs:0.751193080135216,0.00149365197908887);
\draw[draw=black,fill=steelblue31119180,fill opacity=0.75] (axis cs:0.751193080135216,0) rectangle (axis cs:0.800954464108173,0.00373412994772218);
\draw[draw=black,fill=steelblue31119180,fill opacity=0.75] (axis cs:0.800954464108173,0) rectangle (axis cs:0.85071584808113,0.00392083644510829);
\draw[draw=black,fill=steelblue31119180,fill opacity=0.75] (axis cs:0.850715848081129,0) rectangle (axis cs:0.900477232054086,0.00896191187453323);
\draw[draw=black,fill=steelblue31119180,fill opacity=0.75] (axis cs:0.900477232054086,0) rectangle (axis cs:0.950238616027043,0.0125093353248693);
\draw[draw=black,fill=steelblue31119180,fill opacity=0.75] (axis cs:0.950238616027043,0) rectangle (axis cs:1,0.0158700522778193);
\end{axis}

\end{tikzpicture}}
        \caption{}
        \label{fig:subfig2_Startup}
    \end{subfigure}
    \caption{Env 2 IoU Probability Mass Function (PMF). (a) IoU histogram of RC SceneSense predictions. (b) IoU histogram of FC SceneSense predictions. The IoU distributions show that RC occupancy predictions are more likely to be similar than FC predictions.}
    \label{fig:pred_PMF}
\end{figure}

We can analyze the number of available voxels for occupancy prediction as the number of unknown voxels in the target area $x_{rm}$ as a percentage of the total observed voxels in $x_{gt}$.  Using the results from environment 2 to evaluate this prediction, we calculate that on average $59.18\%$ of target area voxels are unknown when performing RC occupancy prediction. However, when performing FC prediction this number jumps to $70.79\%$. This result supports the intuitive statement that there are more available (unknown) voxels for prediction around the frontiers of the map than around the robot.

To confirm that the increase in unknown voxels widens the distribution of occupancy prediction we generate a distribution of results by calculating the IoU of each prediction against all other predictions made during the run. The results of this approach as provided in Figure \ref{fig:pred_PMF} show that RC predictions are more likely to be similar, while FC predictions are more likely to be dissimilar with very little overlap. When predictions are all similar, PMM becomes less important for accurate predictions, since OSMM would result in a similar map each time. However PMM is needed at range to achieve reasonable results since it can negotiate the wider distribution of possible occupancy predictions.

We see an expected increase in FID and KID for frontier-centric prediction. Around the robot, more of the space has been directly observed with lidar sensor measurements. This means less unknown voxels are available for generative prediction since SceneSense does not override occupancy values for voxels that are observed occupied or free based on sensor measurements. As a result, the underlying structure of the environment is more similar between predictions and the ground truth. This results in lower FID and KID values for robot-centric. When making predictions at the frontier, more unknown voxels are available, leading to larger variance in the predictions. This can result in a larger difference between the ground truth map and the predicted occupancy generated with SceneSense, leading to larger FID and KID values.

%%%%%%%%%%%%%%%%%%%%%%%%%%%%%%%%%%%%%%%%%%%%%%%%%%%
% EXTENDED EXPLORATION EXPERIMENTS
%%%%%%%%%%%%%%%%%%%%%%%%%%%%%%%%%%%%%%%%%%%%%%%%%%%

\section{Extended Exploration Experiments}
In this section we examine SceneSense as a ``drop-in'' method for autonomous exploration. First we examine key exploration scenarios where SceneSense solves existing map based exploration problems and evaluate the systems performance at autonomously navigating narrow corners. Then, we evaluate the robots exploration performance on full scene explorations, in 2 dissimilar environments. 
\subsection{Solved Robotic Exploration Tasks}

Robotic systems often exhibit poor or unpredictable performance in edge cases, particularly when encountering rare or unforeseen scenarios. Many edge cases in robotics are known, and need special error handling processes to circumvent errors when in these configurations. With the addition of SceneSense to our robotic framework we alleviate the need for special error handling for some common edge cases seen during deployment. In particular, the common sense occupancy filling alleviates cases where robotic systems cannot generate plans over ares in which they cannot directly observe although it would be common sense that a traversable path exists. We provide two examples of configurations where planning failed without the probabilistic map merging and one example where traditional terrain filling methods would fail where SceneSense does not. 

\subsubsection{Start Up}
\label{sect:startup}
The first configuration where the robotic planner fails in during start up. Given the mounting point of the lidar and height of the robot, there are several meters around the platform that cannot be observed directly by the lidar. At startup this results in a large hole around the platform, which the planner sees as untraversable terrain and therefore is unable to form a plan for traversable as shown in Figure \ref{fig:startup-figure}. Algorithmic hole filling methods attempt to solve this problem but these methods result in unexpected failures when traversing new environments and require online tuning for fill size. Furthermore, these methods require specific handling of the ``start'' configuration requiring additional logic for the robot exploration method. A different approach is to add additional depth sensors intended to point directly at what the robot is walking on. However this requires the purchase of additional sensors which can be prohibitively expensive, and would require additional ongoing calibration between the sensor to ensure alignment of measurements. Alternatively, SceneSense is able to fill the hole at startup and immediately begin planning, without additional sensors or special configuration handling. 

\subsubsection{Narrow Hallways and Corners}
\label{sect:hallway_test}
During Traversal, perception systems can miss measurements of necessary information to continue exploration. In our case, due the the high mounting point of our lidar sensor, traversal of narrow hallways can lead to missing ground measurements necessary to continue exploration as shown in figure \ref{fig:hallway_missed}. In some sampling based planners for ground systems, it is not allowed to sample points above open space, and therefore the robot will never traverse down the hallway to continue exploration. SceneSense provides realistic filling in this scenario and allows the planner to continue exploration down the hallway.

In this scenario we can directly measure the impact that SceneSense has on robot navigation. We provide 2 experiments to evaluate the SceneSense occupancy enhancements when traversing narrow corners. First, we run 10 corner traversals in the same location to evaluate consonant of the enhancements. Second, we select 10 different narrow hallway scenarios to evaluate the SceneSense predictions in different environments and generalization abilities. In both scenarios we set a start and end point during the hallway exploration and set the robot to explore. We evaluate the following systems for comparison.

\begin{enumerate}
    \item \textbf{User Operation Ground Truth}: The system is teleoperated by the user to generate a ground truth result for human-like exploration. 
    \item \textbf{GBPlanner \cite{dang2020graph}}: Baseline GBPlanner is used to explore the narrow hallway environments. 
    \item \textbf{GBPlanner + Local SceneSense }: Baseline GBPlanner is used to explore the narrow hallway environments. SceneSense predictions are merged with the running occupancy map as the system explores. SceneSense only makes local occupancy predictions (Predictions centered around the robot). 
    \item \textbf{GBPlanner + Frontier SceneSense }: Baseline Planner is used to explore the narrow hallway environments. SceneSense predictions are merged with the running occupancy map as the system explores. SceneSense makes predictions at range for 1 target point in the graph with the highest exploration gain. 
\begin{table}[h]
\footnotesize % Ensures the table fits in a single column
\centering
\caption{\textbf{Single Hallway Results}: Single hallway traversal results averaged over 8 runs per configuration. All time measurements are in seconds, and distance measurements are in meters.}
\label{table:hallway_traversal_results}

\begin{tabular}{p{2.2cm} | p{0.75cm} p{0.75cm} p{0.75cm} p{0.75cm} p{0.90cm}} 
\toprule
\textbf{Method} & \textbf{Exp. Time} (s) & \textbf{Min Time} (s) & \textbf{Max Time} (s) & $\boldsymbol{\sigma_{\text{exp}}}$ (s) & \textbf{Fail Count} \\
\midrule
Human Operator & 9.35  & 9.00  & 9.70  & 0.34  & 0 \\ 
GB~\cite{dang2020graph}  & 25.98 & 18.33 & 33.77 & 5.18  & 1 \\
GB + Local SS  & 23.77 & 18.48 & 29.27 & 3.81  & 0 \\
GB + Frontier SS  & 21.10 & 16.35 & 28.17 & 3.53  & 0 \\
\botrule
\end{tabular}
\end{table}

\end{enumerate}

\subsubsection{Discussion}
As shown in Table \ref{table:hallway_traversal_results}, enhancing the running occupancy map using the frontier SceneSense predictions results in the best autonomous performance of the system. The performance of GBPlanner + Frontier SS is not only the lowest mean exploration time and path length, but also has a lower variance than the other methods. We believe that the primary reason that frontier SS outperforms local SS is due to system latency. In particular the latency introduced in running multiple diffusion and merging those predictions into the map. As shown in Figure \ref{fig:Narrow Hallway Predictions} the robot is able to perform 4 consecutive occupancy predictions while traversing the calculated path. Each of these predictions takes approximately $0.5$ seconds to generate and similar time to merge with the running map. That means that by the time that the robot has made it to the area where it will need to plan its next path the SceneSense predictions will have already been merged into the map for quick planning. When performing local occupancy prediction its takes time to to make the occupancy predictions and update the running occupancy map, resulting in the robot frequently pausing at the end of its initial path.  If this prediction and update latency was reduced the local SceneSense predictions should perform just as well as the frontier predictions.

Importantly, a traversability failure is observed in this experiment when operating over the sensor only map. This failure is observed when the robot traverses particularly close to the right wall of the intersection, and therefore cannot observe the ground necessary for traversal. A similar observation problem is show in Figure \ref{fig:hallway_traversal}. The SceneSense enhanced map is able to make this logical inference of ground existence every time in our testing, allowing for more reliable exploration.

\begin{figure}[]
    \centering
    % First subfigure (Environment 1)
    \begin{subfigure}{0.35\linewidth}
        \centering
        \includegraphics[width=\linewidth]{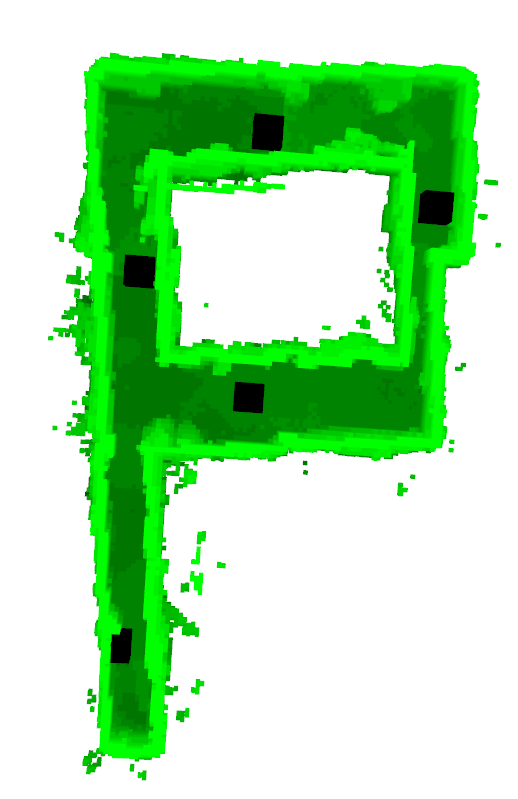}
        \caption{}
        \label{fig:explore_env}
    \end{subfigure}
    \hfill
    % Second subfigure (Environment 2)
    \begin{subfigure}{0.45\linewidth}
        \centering
        \includegraphics[width=\linewidth]{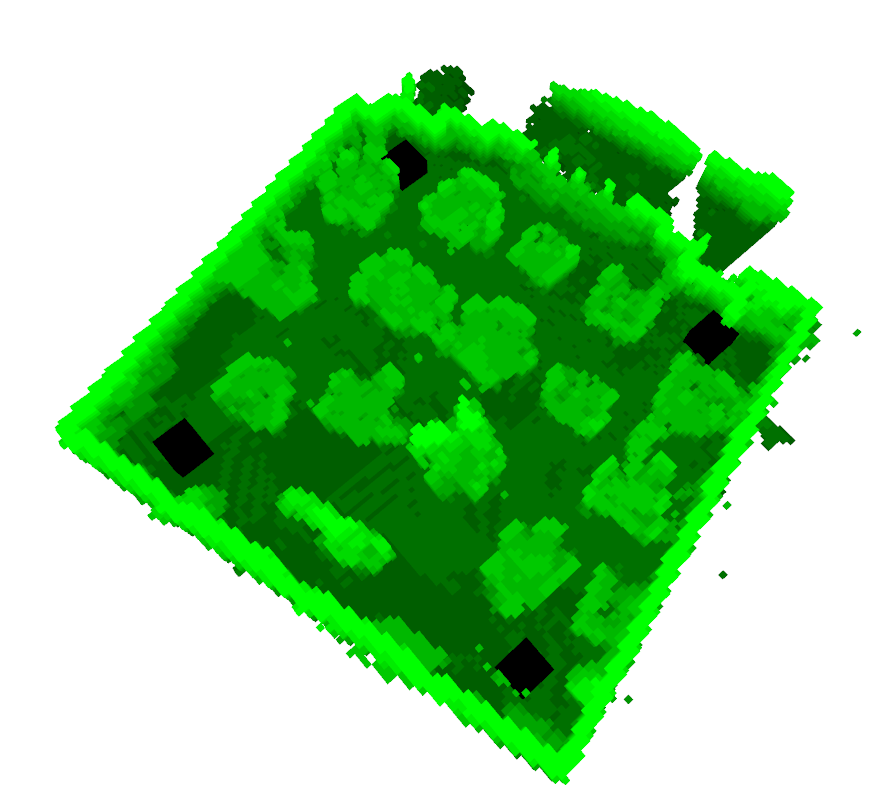}
        \caption{}
        \label{fig:explore_env_2}
    \end{subfigure}

    \caption{Two test environments for exploration evaluation. The green voxels represent the complete occupancy map, and the black cubes indicate the robot's starting positions. (a) Environment 1 (b) Environment 2.}
    \label{fig:combined_env}
\end{figure}

\subsubsection{SceneSense Robustness to Adversarial Scenarios}
While promising, the results presented in Sections \ref{sect:startup} and \ref{sect:hallway_test} could be solved with simple brute force solution such as removing the ground checking from the graph sampler. To further differentiate SceneSense from potential brute force solutions we identify adversarial scenarios where areas of the map are not traversable. One such scenario is shown in Figure \ref{fig:stairs_diff_all}, where the robot traverses near a glass railing unobservable by the lidar sensor. While a naive unconstrained sampling approach could sample points and paths above the empty space, SceneSense is able to differentiate this space from other potential ground areas and not predict ground over the edge.

\subsection{Full Robot Exploration}
In this section we explore the end effects of SceneSense when applied to a robotic exploration platform. For these experiments SceneSense is considered a ``drop-in'' method that does not require custom planning or mapping frameworks. 
\subsubsection{Experimental Setup}

\begin{figure}[t]
    \centering
    % First subfigure
    \begin{subfigure}{0.8\linewidth}
        \centering
        \includegraphics[width=\linewidth]{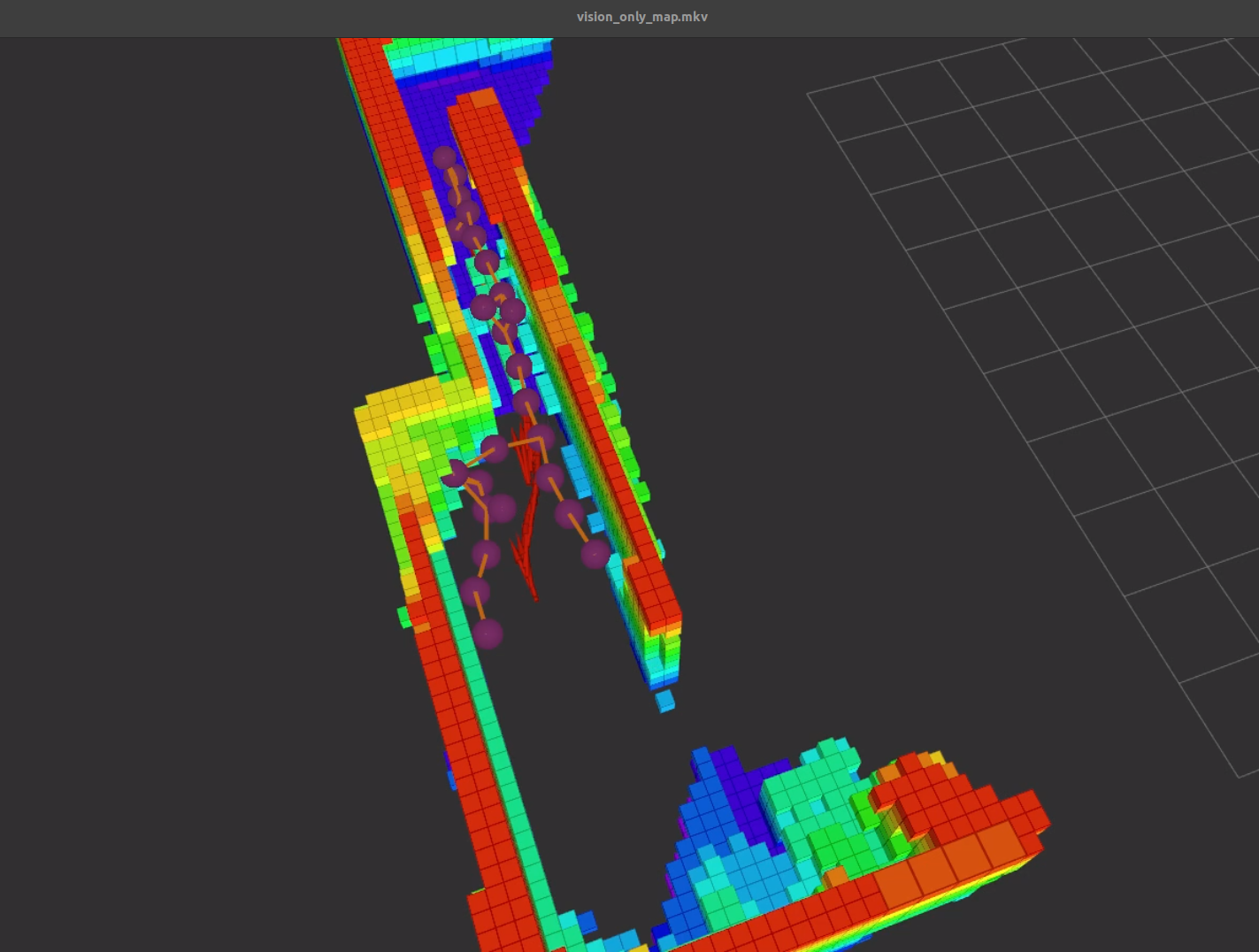}
        \caption{}
        \label{fig:vision_only}
    \end{subfigure}
    \hfill
    % Second subfigure
    \begin{subfigure}{0.8\linewidth}
        \centering
        \includegraphics[width=\linewidth]{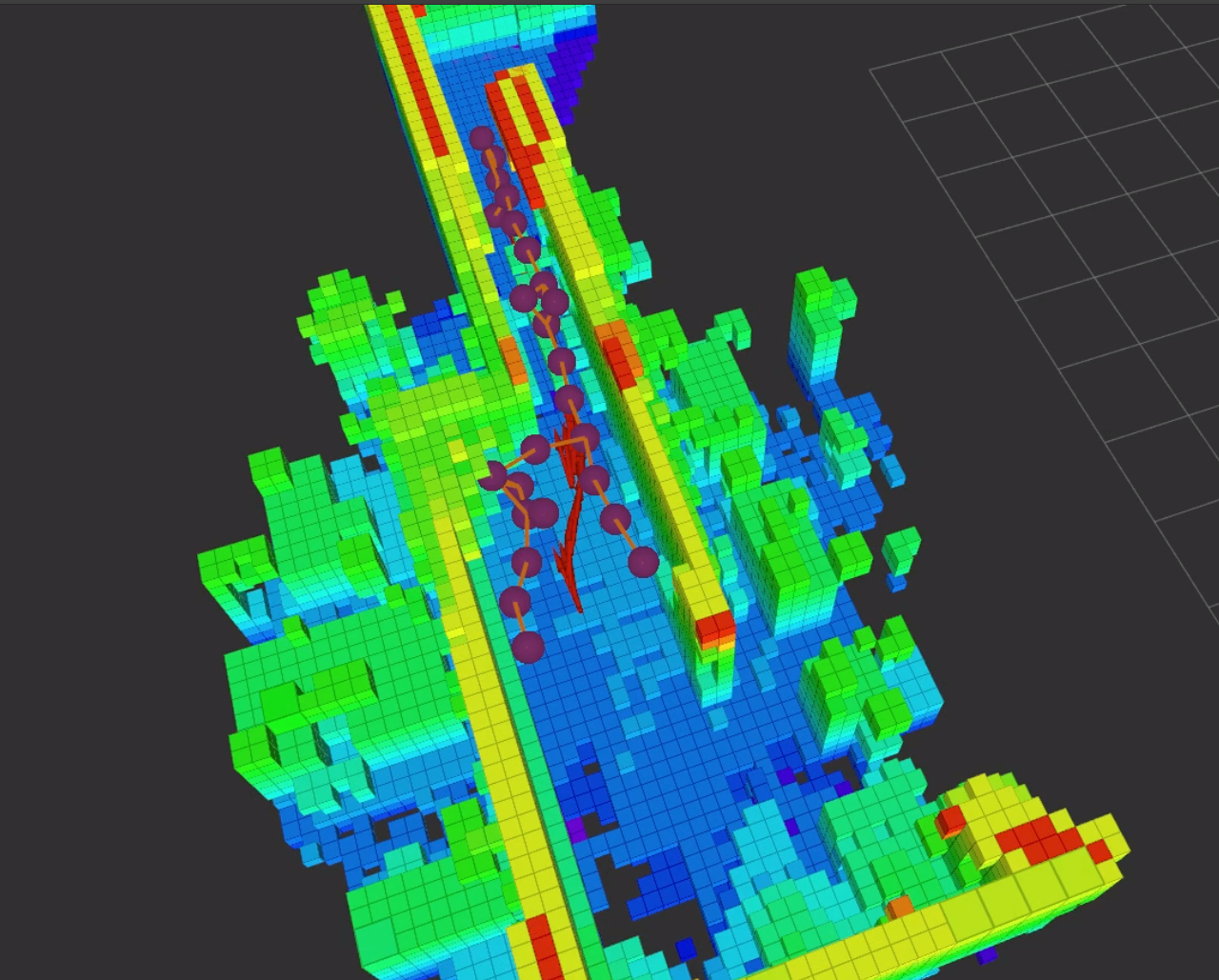}
        \caption{}
        \label{fig:vision_and_ss}
    \end{subfigure}

    \caption{(a) The vision only map. (b) The SceneSense enhanced map. The current planning graph is shown as the purple and red graph structures. In (a), the vision only map contains a large gap near the robot’s starting position. Due to occlusions from the narrow hallway, the hole is not observed and therefore remains unpopulated. SceneSense, however, merges its predictions with the existing map to make a realistic estimation of the unobserved space, filling the holes and allowing the robot to plan over the environment. Traversable occlusions, such as vertical structures, are correctly negotiated out of the map when the robot makes observations of the space.}
    \label{fig:Vision_vs_ss}
\end{figure}

% Horizontally Aligned
% \begin{figure*}[]
%   \centering
%   \hfill
%     \subfigure[]{
%     \includegraphics[width=0.45\textwidth]{figures/teleop_hallway_and_gb_plot_and_ss_and_frontier.png}
%   }
%   \hfill
%     \subfigure[]{
%     \includegraphics[width=0.45\textwidth]{figures/classrooom_all_results_plot.png}
%   }
%   \hfill
%   \caption[Exploration Percentage over Time Env. 1]{\textbf{Exploration Percentage over Time env. 1}: Given 5 total runs per configuration, the robot is started once at each of the starting positions shown in Figure \ref{fig:explore_env}. The average exploration $\%$ is shown for each configuration, where the shaded region is the standard deviation across the 5 runs at that timestep.  Additionally linear approximations are shown to capture the average exploration rate to achieve 95\% exploration of the testing environment.}
%   \label{fig:exploration_plot_1}
% \end{figure*}

\begin{figure*}[]
  \centering
  % First subfigure
  \begin{subfigure}{1.0\textwidth}
    \centering
    \includegraphics[width=\linewidth]{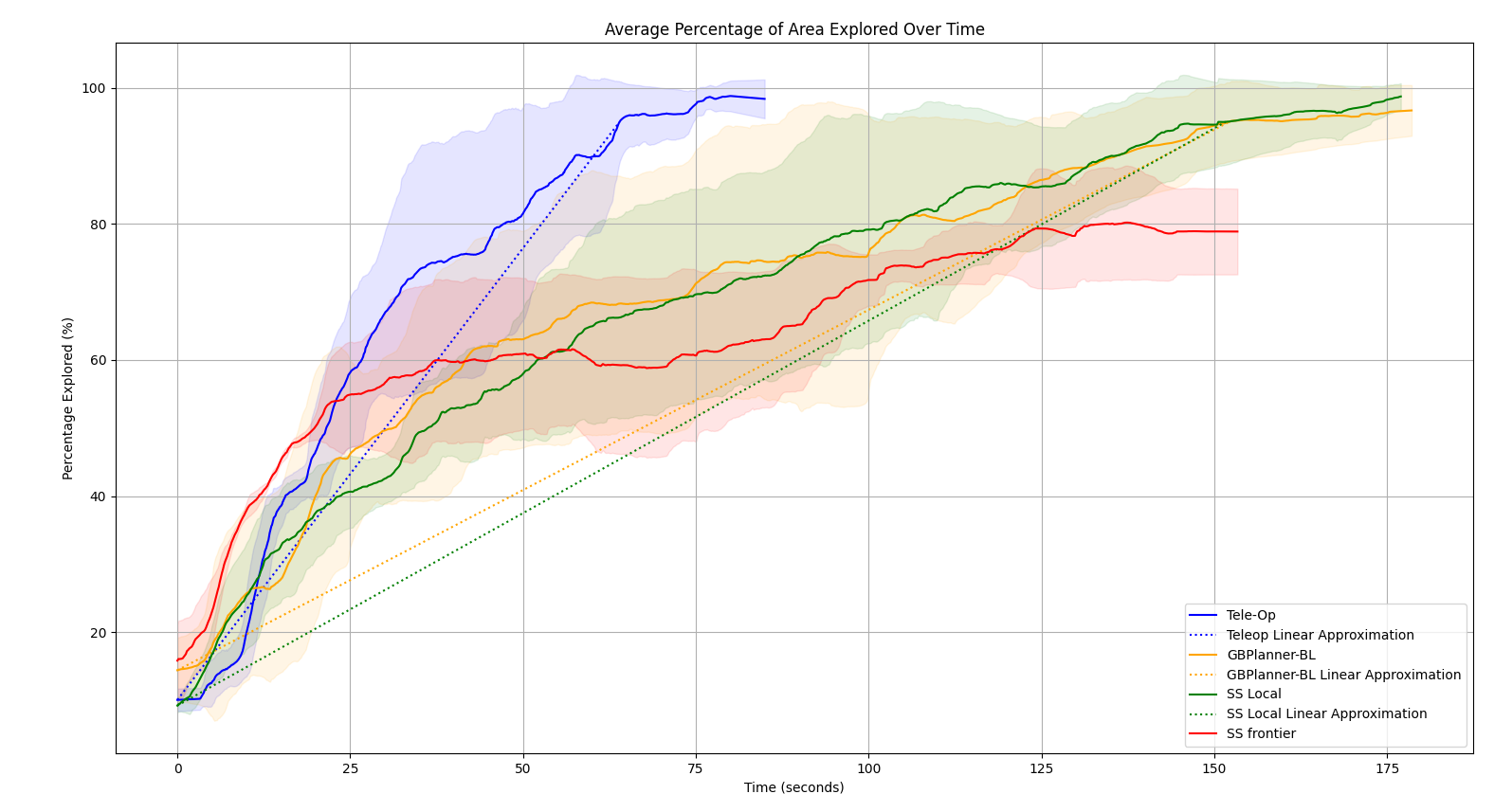}
    \caption{}
    \label{sub-fig:exploration_plot_8th}
  \end{subfigure}
  \hfill
  % Second subfigure
  \begin{subfigure}{1.0\textwidth}
    \centering
    \includegraphics[width=\linewidth]{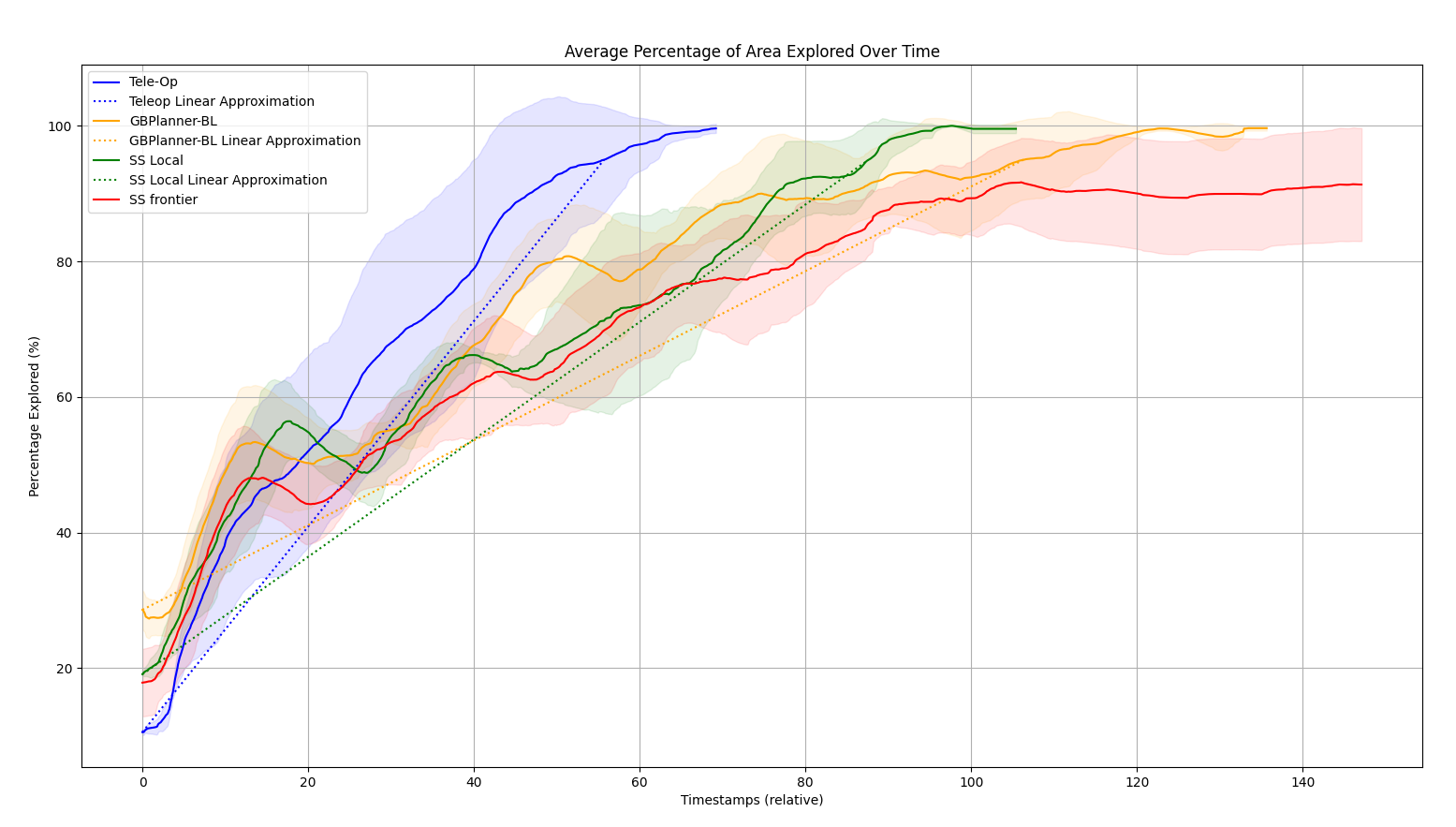}
    \caption{}
    \label{sub-fig:exploration_plot_clsrm}
  \end{subfigure}

  \caption{(a) Results associated with runs in Env. 1, as shown in Figure \ref{fig:explore_env}. (b) Results associated with runs in Env. 2, as shown in Figure \ref{fig:explore_env_2}. The robot starts once at each of the positions in Figures \ref{fig:explore_env} and \ref{fig:explore_env_2}. The average exploration percentage is shown for each configuration, where the shaded region represents the standard deviation across runs at that time step. Additionally, linear approximations illustrate the average exploration rate required to achieve 95\% coverage of the test environment.}
  \label{fig:exploration_plot_1}
\end{figure*}

To evaluate the exploration speed of the system we deployed the platform in test environments shown in Figures \ref{fig:explore_env}and \ref{fig:explore_env_2}. The robot is started once at each position indicated by the black boxes in the associated figures (5 positions in environment 1, 4 positions in environment 2). One run per starting position is then run per configuration, and the results are averaged to generate Figure \ref{fig:exploration_plot_1} and Tables \ref{table:exploration_1} and \ref{table:full_explore_classroom}. The same testing configurations described in Section \ref{sect:hallway_test} are used for all runs.

\begin{table}[h]
\footnotesize % Shrinks text to fit within a single column
\caption{\textbf{Env. 1 Exploration Results}: Results from exploration of the environment shown in Figure~\ref{fig:explore_env}. All results are calculated based on reaching 95\% environment exploration.}
\label{table:exploration_1}
\begin{tabular}{p{2.2cm} | p{0.75cm} p{0.75cm} p{0.75cm} p{0.75cm} p{0.75cm}}
\toprule
\textbf{Method} & \textbf{Exp. Time} (s) & \textbf{Min Time} (s) & \textbf{Max Time} (s) & $\boldsymbol{\sigma_{\text{exp}}}$ (s) & $\boldsymbol{\frac{\text{exp} \%}{\text{sec}}}$ \\
\midrule
Human Operator  & 64.03  & 36.75  & 75.05  & 8.717  & 1.327 \\ 
GB~\cite{dang2020graph}  & 152.27 & 57.75  & 148.45 & 13.027 & 0.529 \\
GB + Local SS  & 151.83 & 91.65  & 141.99 & 9.629  & 0.565 \\
GB + Frontier SS  & N.A.  & N.A.   & N.A.   & 7.86   & N.A.  \\
\botrule
\end{tabular}
\end{table}

% \begin{table}[h]
% \small % Shrinks text to fit within a single column
% \caption{\textbf{Env. 1 Exploration Results}: Results from exploration of the environment shown in Figure~\ref{fig:explore_env}. All results are calculated based on reaching 95\% environment exploration.}
% \label{table:exploration_1}
% \begin{tabular}{|p{1.4cm}|p{0.8cm}|p{0.8cm}|p{0.8cm}|p{0.8cm}|p{0.8cm}|}
% \hline
% \textbf{Method} & \textbf{Exp. Time} & \textbf{Min Time} & \textbf{Max Time} & $\boldsymbol{\sigma_{\text{exp}}}$ & $\boldsymbol{\frac{\text{exp} \%}{\text{sec}}}$ \\
% \hline
% Human Operator  & 64.03  & 36.75  & 75.05  & 8.717  & 1.327 \\ 
% \hline
% GB~\cite{dang2020graph}  & 152.27 & 57.75  & 148.45 & 13.027 & 0.529 \\
% \hline
% GB + Local SS  & 151.83 & 91.65  & 141.99 & 9.629  & 0.565 \\
% \hline
% GB + Frontier SS  & N.A.  & N.A.   & N.A.   & 7.86   & N.A.  \\
% \hline
% \end{tabular}
% \end{table}

\begin{table}[h]
\footnotesize % Ensures it fits in a single column
\caption{\textbf{Env. 2 Exploration Results}: Results from exploration of the environment shown in Figure~\ref{fig:explore_env_2}. All results are calculated based on reaching 95\% environment exploration.}
\label{table:full_explore_classroom}
\begin{tabular}{p{2.2cm} | p{0.75cm} p{0.75cm} p{0.75cm} p{0.75cm} p{0.75cm}}
\toprule
\textbf{Method} & \textbf{Exp. Time} & \textbf{Min Time} & \textbf{Max Time} & $\boldsymbol{\sigma_{\text{exp}}}$ & $\boldsymbol{\frac{\text{exp} \%}{\text{sec}}}$ \\
\midrule
Human Operator  & 55.76  & 40.95  & 62.55  & 10.445 & 1.516 \\ 
GB~\cite{dang2020graph}  & 106.46 & 69.68  & 117.55 & 5.394  & 0.624 \\
GB + Local SS  & 81.68  & 74.35  & 91.05  & 5.685  & 0.866 \\
GB + Frontier SS  & N.A.  & N.A.   & N.A.   & 6.993  & N.A.  \\
\botrule
\end{tabular}
\end{table}

\subsubsection{Discussion}

Our experiments demonstrate that SceneSense-enhanced maps consistently improve exploration performance compared to vision only maps. In both test environments, SceneSense yields lower average and maximum exploration times with higher exploration gain per second. The vision only maps showed faster minimum exploration times, but with much larger maximum exploration times, indicating far greater variance. These results suggest lower consistency of operation when relying solely on vision. In some starting configurations the robot may make observations that enable rapid exploration, while in others it may make poor observations that increase total exploration time. Local occupancy prediction narrows this distribution, resulting in a system that is less dependent on a favorable direct observation at startup.

Our data revealed that startup configurations lead to some environment specific behavior. The minimum time in environment 1 is attributed to the vision-only GBPlanner navigating straight down the hallway when starting from the dead end, instead of turning right at the first junction as seen in Figure \ref{fig:explore_env}. With the more complete ground representation from the SceneSense enhanced map, GBPlanner instead navigates down the larger hallway to the right first, which increases its overall completion time. Without this outlier in the environment 2 data collection, the exploration rate of the SceneSense enhanced planner increases by 32\%. It is our hypothesis that the substantial exploration rate gains are actually attributed to the startup filling. As shown in Figure \ref{fig:vision_and_ss}, the startup filling has a lingering effect on the map even after some initial exploration. It is likely that the exploration gain in these areas remains significant, resulting in the planner choosing to double back to explore the starting position in several instances.

\begin{figure}
    \centering
    \includegraphics[width=1.0\linewidth]{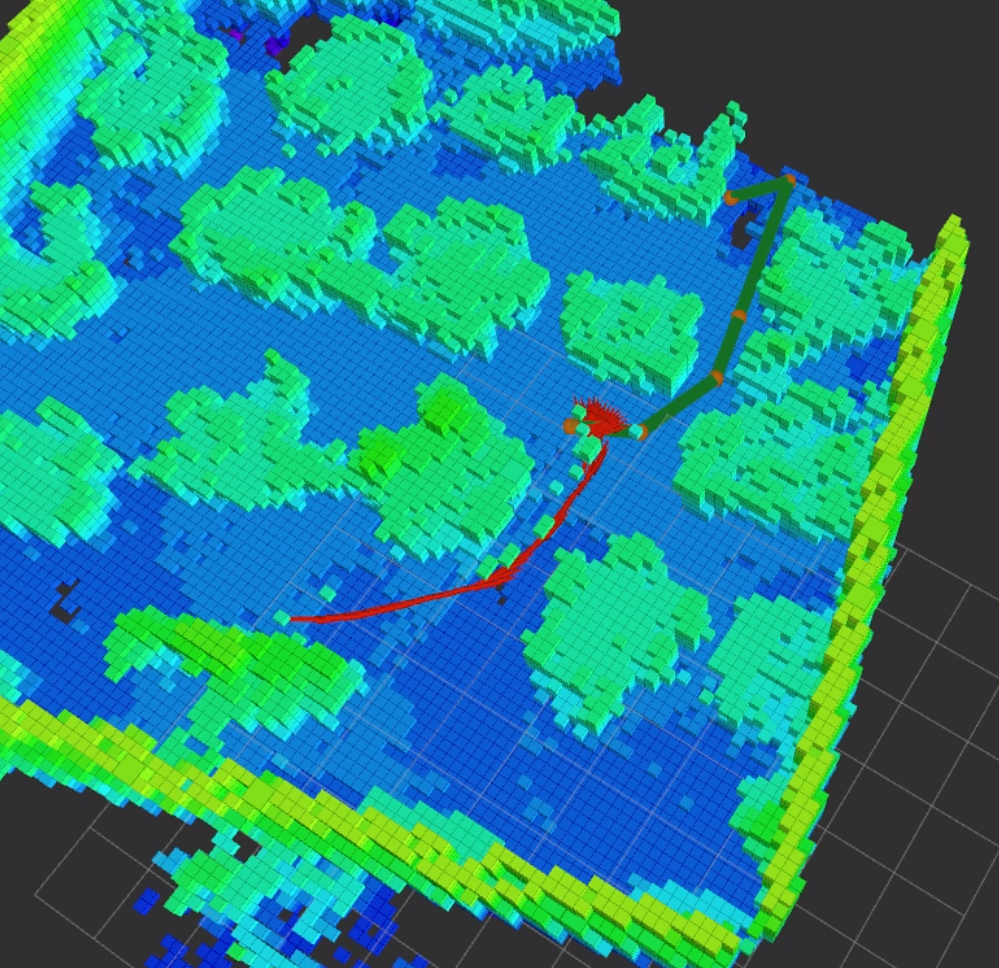}
    \caption{\textbf{Unintuitive Planning.} As the robot is traversing the classroom environment, a hole in the map leads to a loop-back path. 
    The robot odometry is shown in red, and the planned path is shown in green. 
    Due to the hole in the map, the robot's plan configures it into a poor position to continue exploration.}
    \label{fig:Unintuitive_planning}
\end{figure}

We observed a clear distinction between local SceneSense predictions and frontier-centric SceneSense predictions. Local predictions improved planning performance both by filling in startup regions and by eliminating small holes in the map that otherwise can cause unintuitive behavior. For instance, in Figure \ref{fig:Unintuitive_planning}, a small hole causes the planner to loop back on itself rather than taking a direct path. SceneSense fills such gaps, producing maps that enable coherent trajectory generation and thereby contributing to faster and more consistent exploration.

In contrast, frontier-centric SceneSense (SS frontier), while initially promising, was unable to complete exploration as seen in Figure \ref{fig:exploration_plot_1}. This failure arises because diffusion predictions overwrite unobserved voxels with occupancy probability values, which erases the boundary between free and unknown that frontier planners rely on. As a result, the planner prematurely concludes that the environment is fully explored. One possible mitigation is to add an indicator variable that marks whether a voxel was generated by diffusion or observed directly. While this allows for differentiation between diffused predicted voxels and sensor observed voxels, it does not recover the geometry of the frontier and would require the planner to inherently understand when to reason over flag based voxels and when to reason over the traditional sensor observed voxel representations. In practice, the planner would need to reason explicitly over the interface between diffusion-generated occupancies and those derived from sensor measurements in order to continue exploration using frontier based exploration. These results suggest that new planners are needed which can reason jointly about observed and predicted occupancy, preserving the structural cues necessary for frontier-based exploration.

Although SS frontier did not complete exploration, we argue it remains a valid and motivating contribution. First, it demonstrates the potential of extending SceneSense beyond local completions toward map-wide inference. Second, its failure case highlights the broader need for exploration planning paradigms that explicitly incorporate predictive models with joint reasoning, rather than treating predictions as direct drop in substitutions for perception. These insights point toward important research questions on integrating predicted occupancy into higher level reasoning and planning systems, which pursue in future work.

%%%%%%%%%%%%%%%%%%%%%%%%%%%%
% Limitations
%%%%%%%%%%%%%%%%%%%%%%%%%%%%
\section{Limitations}\label{sec:limitations}
Despite the promising results demonstrated by our proposed method for generative occupancy map synthesis, several limitations warrant discussion and present opportunities for future work.

Currently, training and evaluation of SceneSense (SS) is limited to structured indoor environments. The generalization of our method to more diverse, complex scenarios, such as outdoor, unstructured, or highly dynamic environments, remains an open question. The training data, collected from a single building, may not fully capture the variability present in real-world conditions. Future work could train the model on broader datasets to improve its robustness and generalizability.

Our system's reliance on an off-board GPU for inference presents a practical limitation for autonomous systems with strict power and size constraints. Deployment on embedded platforms would require model optimization and potentially an exploration of more efficient architectures.

Incorrect predictions, particularly in scenarios involving transparent or reflective surfaces, could produce unpredictable behavior. While the probabilistic map merging (PMM) framework helps mitigate the impact of some erroneous predictions by favoring sensor measurements, a single, highly confident but incorrect prediction can still adversely affect the map's quality and the robot's subsequent planning.

Finally, while our approach improved map quality and traversability, it did not always result in a complete exploration of the environment, as seen in the case of the SS-frontier approach (Figure 13). This suggests a potential mismatch between the current frontier detection algorithms, which may be confused by the predicted occupancy values, and the generative map. Future work addressing frontier exploration methods for generative occupancy could ensure that the robot can successfully complete exploration.

%%%%%%%%%%%%%%%%%%%%%%%%%%%%
% CONCLUSION
%%%%%%%%%%%%%%%%%%%%%%%%%%%%

\section{Conclusion}\label{sec:conclusion}

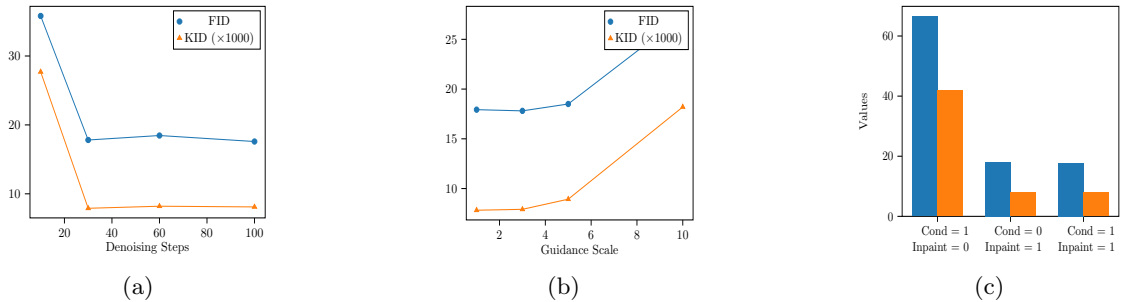
\begin{figure*}[t]
    \centering
    % First subfigure
    \begin{subfigure}{0.3\textwidth}
        \centering
        \resizebox{100pt}{95pt}{% This file was created with tikzplotlib v0.10.1.
\begin{tikzpicture}

\definecolor{darkgray176}{RGB}{176,176,176}
\definecolor{darkorange25512714}{RGB}{255,127,14}
\definecolor{steelblue31119180}{RGB}{31,119,180}

\begin{axis}[
tick align=outside,
tick pos=left,
x grid style={darkgray176},
xlabel={Denoising Steps},
xmin=5.5, xmax=104.5,
xtick style={color=black},
y grid style={darkgray176},
ymin=6.5045, ymax=37.2055,
ytick style={color=black}
]
\addplot [draw=steelblue31119180, fill=steelblue31119180, mark=*, only marks]
table{%
x  y
10 35.81
30 17.81
60 18.46
100 17.59
};
\addplot [draw=darkorange25512714, fill=darkorange25512714, mark=triangle*, only marks]
table{%
x  y
10 27.7
30 7.9
60 8.2
100 8.1
};
\addplot [semithick, steelblue31119180]
table {%
10 35.81
30 17.81
60 18.46
100 17.59
};
\addplot [semithick, darkorange25512714]
table {%
10 27.7
30 7.9
60 8.2
100 8.1
};
\addlegendentry{FID}
\addlegendentry{KID ($\times 1000$)}

\end{axis}

\end{tikzpicture}}
        \caption{}
        \label{fig:denoising_steps}
    \end{subfigure}
    \hfill
    % Second subfigure
    \begin{subfigure}{0.3\textwidth}
        \centering
        \resizebox{100pt}{95pt}{% This file was created with tikzplotlib v0.10.1.
\definecolor{darkgray176}{RGB}{176,176,176}
\definecolor{darkorange25512714}{RGB}{255,127,14}
\definecolor{steelblue31119180}{RGB}{31,119,180}

\begin{tikzpicture}
\begin{axis}[
tick align=outside,
tick pos=left,
x grid style={darkgray176},
xlabel={Guidance Scale},
xmin=0.55, xmax=10.45,
xtick style={color=black},
y grid style={darkgray176},
ymin=6.844, ymax=28.316,
ytick style={color=black}
]
\addplot [draw=steelblue31119180, fill=steelblue31119180, mark=*, only marks]
table{%
x  y
1 17.93
3 17.81
5 18.5
10 27.34
};
\addplot [draw=darkorange25512714, fill=darkorange25512714, mark=triangle*, only marks]
table{%
x  y
1 7.82
3 7.91
5 8.93
10 18.2
};
\addplot [semithick, steelblue31119180]
table {%
1 17.93
3 17.81
5 18.5
10 27.34
};
\addplot [semithick, darkorange25512714]
table {%
1 7.82
3 7.91
5 8.93
10 18.2
};

\addlegendentry{FID}
\addlegendentry{KID ($\times 1000$)}

\end{axis}

\end{tikzpicture}}
        \caption{}
        \label{fig:guidance_scale}
    \end{subfigure}
    \hfill
    % Third subfigure
    \begin{subfigure}{0.3\textwidth}
        \centering
        \resizebox{100pt}{95pt}{% This file was created with tikzplotlib v0.10.1.
\begin{tikzpicture}

\definecolor{darkgray176}{RGB}{176,176,176}
\definecolor{darkorange25512714}{RGB}{255,127,14}
\definecolor{steelblue31119180}{RGB}{31,119,180}

\begin{axis}[
tick align=outside,
tick pos=left,
x grid style={darkgray176},
xmin=-0.485, xmax=2.485,
xtick style={color=black},
xtick={0,1,2},
x tick label style={align=right, text width=2cm},
xticklabels={Cond = 1 
Inpaint = 0,Cond = 0
Inpaint = 1,Cond = 1
Inpaint = 1},
y grid style={darkgray176},
ylabel={Values},
ymin=0, ymax=69.867,
ytick style={color=black}
]
\draw[draw=none,fill=steelblue31119180] (axis cs:-0.35,0) rectangle (axis cs:0,66.54);
% \addlegendimage{ybar,ybar legend,draw=none,fill=steelblue31119180}
% \addlegendentry{FID}
\draw[draw=none,fill=steelblue31119180] (axis cs:0.65,0) rectangle (axis cs:1,17.98);
\draw[draw=none,fill=steelblue31119180] (axis cs:1.65,0) rectangle (axis cs:2,17.81);
\draw[draw=none,fill=darkorange25512714] (axis cs:2.77555756156289e-17,0) rectangle (axis cs:0.35,41.85);
\addlegendimage{ybar,ybar legend,draw=none,fill=darkorange25512714}
\draw[draw=none,fill=darkorange25512714] (axis cs:1,0) rectangle (axis cs:1.35,7.94);
\draw[draw=none,fill=darkorange25512714] (axis cs:2,0) rectangle (axis cs:2.35,7.91);
\end{axis}

\end{tikzpicture}}
        \caption{}
        \label{fig:bar_chart}
    \end{subfigure}

    \caption{\textbf{SceneSense Ablation Experiments.} All ablation experiments were conducted on Test House 1 using the same trained diffusion network. For all experiments, conditioning and inpainting were enabled, $s$ was set to 3, and 30 denoising steps were used unless these values were being ablated. (a) Evaluation of various denoising step values. (b) Evaluation of different guidance scale $s$ values, further discussed in \ref{sect:guidance-scale}. (c) Ablation of conditioning and inpainting for the network, where 1 indicates enabled and 0 indicates disabled.}
    \label{fig:ablations}
\end{figure*}

In this work we have presented the general occupancy mapping tool, SceneSense, implemented the model onboard a real-world system, and evaluated both the realism of the predictions as well as the predictions impacts on robotic exploration. We show that the SceneSense model is able to enhance existing occupancy mapping methods and use those hybrid maps as a basis for more constant exploration of space. Further we show that the integration of SceneSense into existing planning frameworks enables the planner to traverse environments that are traditionally challenging to complete with only direct observations, such as narrow hallway traversal or directly after robot startup. Future work includes the design and implementation of planners that can distinguish predicted and observed space, and balance decision making based on observations made in both modalities. Furthermore the SceneSense model can be retrained and scaled up to account for more diverse environments using readily available online datasets in a foundation model style approach. We present SceneSense as an empirically validated approach for enhancing existing robotic planning stacks by providing a ``drop-in'' method that allows for robots to make common-sense inference of unobserved geometry. 

\begin{appendices}

\section*{Appendix A - SceneSense Ablation Study}
\label{sect:app_a}
This ablation study was first presented in \cite{reed2024scenesense} where the SceneSense occupancy prediction network was a conditional diffusion model (as opposed to the unconditional model presented here). When examining these results it was concluded that in this particular application the conditional network does not provided substantial benefits over the equivalent unconditional network. 

\subsection*{Denoising Steps Discussion}
The number of diffusion steps defines the size of each diffusion step during the reverse diffusion process. Generating reasonable results using the fewest possible denoising steps is desirable behavior to reduce computation time. Additionally too many denoising steps have been shown to introduce sampling drift which results in decreased performance \cite{chen2022diffusiondet, ji2023ddp}. As shown in  \ref{fig:ablations} (a) our method saw the best results when configured to 30 denoising steps. Too few denoising steps results in the network being unable make accurate predictions over the large time step. Increasing the number of denoising steps keeps results relatively stable over time, however you can see the KID is slightly worse using more steps due to sampling drift. Sampling drift is a result of the discrepancy between the distribution of the training and the inference data. During training, the model is trained to reduce a noisy map $x_t$ to a ground truth map $x$, at inference time the model iteratively removes noise from its already imperfect noise predictions. These predictions will drift away from the initial corruption distribution which becomes more pronounced at smaller time steps due to the compounding error.

\subsection*{Conditioning and Guidance Scale Discussion}\label{sect:guidance-scale}
The guidance scale $s$ as defined in \cite{ho2020denoising} is a constant that multiplies the difference of the conditional diffusion and unconditional diffusion to ``push'' the diffusion process towards the conditioned answer. Setting $s$ too high results in too large of pushes away from reasonable predictions and results in poor generalization to new environments. The best FID is measured when $s = 3$ however KID is slightly lower when $s = 1$. Further, when examining chart (c) it is shown that the results when conditioning is removed all together ($s = 0$ ) are very similar to the best results seen with conditioning enabled (albeit slightly worse). This is likely because most of the useful conditioning information is captured in the local occupancy data, and mapping measured RGB-D points from areas in front of the geometry to under or behind the platform is a very difficult task. The performance of the conditioning only data may be seen to have a larger impact on the overall results if the sensor could capture more local information, such as given a wider FOV or different mounting angle. 

\section*{Appendix B - Conditional VS. Unconditional Diffusion}
\label{sect:app_b}
This ablation study was first presented in \cite{reed2024online} to evaluate the inference time gains of removing cross-attention based conditioning from the SceneSense occupancy prediction network. 

In this ablation we implement the same denoising network structure presented in \cite{reed2024scenesense}. It is a U-net constructed from the HuggingFace Diffusers library of blocks \cite{von-platen-etal-2022-diffusers} and consists of Resnet \cite{he2015resnet} downsampling/upsampling blocks. The diffusion model is trained using randomly shuffled pairs of ground truth local occupancy maps $x$. We use Chameleon cloud computing resources \cite{keahey2020chameleon} to train our model on one A100 with a batch size of $32$ for $250$ epochs or $88,208$ training steps. We use a cosine learning rate scheduler with a 500 step warm up from $10^{-6}$ to $10^{-4}$. The noise scheduler for diffusion is set to $1,000$ noise steps. 

At inference time we evaluate our dataset using an RTX 4070 TI Super GPU for acceleration. The number of diffusion steps is configured to $30$ steps.
\subsection*{Inference time}
 We evaluate the inference time of the unconditional diffusion model against the inference time of the conditional model presented in the original SceneSense paper \cite{reed2024scenesense}. The cross-attention enabling trainable parameters are removed for the unconditional model, but the number of output channels for the constructed U-net are held constant between both models. As the ablation results of the original paper show minor, or no performance gain between the conditional and unconditional model in this configuration we do not evaluate the results of the model predictions in these experiments. 
\begin{table}[h]
\small % Shrinks text to fit within a single column
\centering
\caption{Inference time and model size results. ``Full inference'' and ``end-to-end'' evaluations are computed using 30 diffusion steps.}
\label{table:inference_time}

\renewcommand{\arraystretch}{1.2} % Adjust row spacing
\setlength{\tabcolsep}{4pt} % Adjust column spacing to fit in one column

\begin{tabular}{p{2.9cm} | p{1.9cm} p{1.9cm}} 
\toprule
& \textbf{Cond. Model}~\cite{reed2024scenesense} & \textbf{Uncond. Model} \\
\midrule
\textbf{Trainable Params} & 141,125,261 & \textbf{101,144,845} \\
\textbf{Diffusion Step (s)} & 0.03707 & \textbf{0.0147} \\
\textbf{Full Inference (s)} & 1.11 & \textbf{0.4437} \\
\textbf{Backbone (s)} & 0.55099 & N.A. \\
\textbf{End-to-End (s)} & 1.66 & \textbf{0.4437} \\
\botrule
\end{tabular}
\end{table}

\vspace{5pt}\\
As shown in Table \ref{table:inference_time}, 
%and as hypothesized by the SceneSense authors \cite{reed2024scenesense} 
removing the conditioning from the diffusion model reduces the computation requirements substantially. The unconditioned model reduces the number of trainable parameters by $28\%$, the model inference time by $60\%$ and the end to end computation time by $73\%$. These improvements enable SceneSense to operate in real-time more effectively, allowing for more flexible implementations for onboard robotic applications.

\end{appendices}


\bibliography{my_bib}% common bib file

\end{document}